\documentclass[11pt]{article}

\usepackage[preprint]{acl}

\usepackage{times}
\usepackage{latexsym}

\usepackage[T1]{fontenc}
\usepackage[utf8]{inputenc}

\usepackage{microtype}
\usepackage{enumitem}
\usepackage{inconsolata}

\usepackage{graphicx}

\usepackage{amsmath}
\usepackage{amsfonts}
\usepackage[T1]{fontenc}
\usepackage[most]{tcolorbox}
\usepackage{url}
\usepackage{booktabs}
\usepackage[normalem]{ulem}
\useunder{\uline}{\ul}{}

\usepackage{amssymb}% http://ctan.org/pkg/amssymb
\usepackage{pifont}% http://ctan.org/pkg/pifont
\title{Protecting Language Models Against Unauthorized Distillation through Trace Rewriting}

\author{Xinhang Ma, William Yeoh, Ning Zhang, Yevgeniy Vorobeychik \\
        Washington University in St. Louis \\
        \texttt{\{m.owen, wyeoh, zhang.ning, yvorobeychik\}@wustl.edu}
    }

\begin{document}
\maketitle

\begin{abstract}
Knowledge distillation is a widely adopted technique for transferring capabilities from LLMs to smaller, more efficient student models.
However, unauthorized use of knowledge distillation takes unfair advantage of the considerable effort and cost put into developing frontier models.
We investigate methods for modifying teacher-generated reasoning traces to achieve two objectives that deter unauthorized distillation: 
(1) \emph{anti-distillation}, or degrading the training usefulness of query responses, 
and (2) \emph{API watermarking}, which embeds verifiable signatures in student models.
We introduce several approaches for dynamically rewriting a teacher's reasoning outputs while preserving answer correctness and semantic coherence.
Two of these leverage the rewriting capabilities of LLMs, while others use gradient-based techniques.
Our experiments show that a simple instruction-based rewriting approach achieves a strong anti-distillation effect while maintaining or even improving teacher performance.
Furthermore, we show that our rewriting approach also enables embedding watermarks that can be reliably detected
%highly reliable watermark detection 
with essentially no false alarms.
Our code is available at \url{https://github.com/xhOwenMa/trace-rewriting}.
\end{abstract}

\section{Introduction}\label{sec:intro}

Knowledge distillation is a simple learning technique for transferring knowledge from one model (the \emph{teacher}) to another (the \emph{student})~\cite{hinton2015distilling}.
In its simplest form, distillation proceeds by querying a teacher model $\mathcal{T}$ with inputs $x$ to obtain responses $y = \mathcal{T}(x)$, then training a student model $\mathcal{S}$ on the resulting input-output pairs using supervised fine-tuning.
Since its introduction, knowledge distillation has become a major workhorse in machine learning across a broad array of applications~\cite{gou2021knowledge, sanh2019distilbert}, as it enables using large, complex, and expensive to execute teachers to train smaller and more efficient student models.
These student models can then be deployed at lower cost 
%(e.g., requiring cheaper computational platforms) 
and exhibit lower inference time, which can be a critical enabler in real-time applications.

However, the simplicity and effectiveness of knowledge distillation even with only a black-box query access to the teacher $\mathcal{T}$---for example, when $\mathcal{T}$ is proprietary---means that this technique can also be used for knowledge and capability ``stealing''.
This issue is especially acute with frontier reasoning-capable LLMs, the design and training of which carry enormous effort and expense~\cite{guo2025deepseek, jaech2024openai}.
These models produce explicit reasoning traces---structured outputs that decompose problem-solving into intermediate steps before arriving at a final answer.
These traces provide rich supervision signals that go beyond mere input-output pairs.
As a result, practitioners increasingly seek to distill reasoning capabilities from these frontier models.
Since distillation incurs a fraction of the cost of training an original teacher, its efficacy without any guardrails can disincentivize innovation.

Two classes of approaches have been proposed to counter unauthorized distillation of large language models: \emph{anti-distillation} and \emph{API watermarking}. 
Anti-distillation methods aim to degrade the efficacy of distillation training~\cite{li2025doge,savani2025antidistillation}.
However, state-of-the-art approaches for anti-distillation significantly degrade the \emph{teacher} efficacy as well as the student, making them impractical.
API watermarking, on the other hand, attempts to insert a watermark into the query responses in a way that enables its verification from the student model trained on such traces~\cite{he2022protecting,he2022cater,zhao2022distillation,zhao2023protecting}.
However, most API watermarking approaches use token-level statistics, and tend to result in a non-negligible false-alarm rate which provides plausible deniability for the model ``thief''.

%While extensive research has focused on improving distillation effectiveness from the student's perspective~\cite{gu2023minillm,mukherjee2023orca}, comparatively little attention has been paid to how \emph{teacher model providers} can influence the distillation process.
We propose and analyze several methods for modifying teacher-generated reasoning traces to achieve these two complementary objectives.
Specifically, we propose two classes of approaches for anti-distillation.
The first uses an assistant LLM to rewrite instructions specifically in order to subvert their use in downstream training while preserving semantics.
This approach leverages the semantic understanding of modern LLMs to transform clean traces into modified versions that achieve our objectives.
Additionally, we adapt our instruction rewriting approach to facilitate stealthy watermark embedding in reasoning traces.
Our second approach uses (projected) gradient-based optimization in the embedding space with the explicit objective of degrading training efficacy of a (proxy) student.
%The first is \textbf{anti-distillation}: degrading the learning efficacy of traces to discourage unauthorized distillation while maintaining the teacher's utility.
%The second is \textbf{output watermarking}: embedding signatures into reasoning traces such that student models trained on these traces inherit detectable characteristics.

Conceptually, the problem of anti-distillation is closely related to the extensive literature on poisoning attacks in machine learning~\cite{goldblum2022dataset,tian2022comprehensive,vorobeychik2018adversarial}, and large language models in particular~\cite{wan2023poisoning,das2025security}.
In this literature, an array of threat models has been considered, including attacks that modify the inputs $x$, the labels $y$, or both.
However, our setting is distinct from nearly all prior work on data poisoning in four important ways.
First, we assume the teacher model is queried sequentially, meaning \emph{we do not have access to the full dataset} in modifying the response to each query, but must instead do so online for each query.
Second, our modification of responses $y$ to queries $x$ must preserve a high degree of functionality to avoid degrading the teacher model.
Third, our responses must be modified in a way that is \emph{stealthy}, ruling out addition of non-nonsensical tokens in obvious positions, such as at the end of the normal response, where they can be easily detected and removed by an adaptive student.
Fourth, the rich space of LLM responses $y$ allows far more opportunities for manipulation than the typical label modification attacks.

We evaluate our methods on LLM reasoning benchmarks using a variety of student model architectures.
Our results show that our optimized instruction-based rewriting approach achieves strong anti-distillation effects, reducing student accuracy by up to 61.3\%---a significantly stronger anti-distillation effect than the recent baseline approaches.
At the same time, our approach maintains and often even \emph{improves} teacher performance, in contrast to baselines, which exhibit significant teacher performance degradation.
We also observe a scaling property where stronger student models experience greater performance degradation, suggesting that capable models more effectively learn the corrupted reasoning patterns.
For API watermarking, our approach enables the embedding of watermarks into student models that can be reliably detected with few verification query while attaining \emph{an essentially zero false alarm rate}, significantly outperforming state-of-the-art API watermarking baselines.

In summary, our main contributions are:
\begin{enumerate}[leftmargin=*,topsep=0pt,itemsep=-1ex]
% \begin{enumerate}
    \item Several prompt-based and gradient-based rewriting approaches for anti-distillation.

    \item A prompt-based rewriting approach for stealthy watermark embedding.
    
    \item Extensive experiments demonstrating that our rewriting approaches achieve (a) state-of-the-art anti-distillation effectiveness without compromising teacher accuracy, and (b) state-of-the-art watermarking reliability with essentially zero false alarms.
    %---outperforming both sampling-based and post-training baselines.

    %\item Extensive experiments demonstrating that our watermarking approach is significantly more reliable (with a near-perfect verification rate and essentially zero false alarm rate) than state-of-the-art baselines.
\end{enumerate}

\section{Related Work}\label{sec:related_work}

\iffalse
\noindent\textbf{Knowledge Distillation:}
%\label{sec:related_work:kd}
Knowledge distillation (KD)~\cite{hinton2015distilling} transfers capabilities from large language models to smaller student models~\cite{xu2024survey}.
Supervised fine-tuning on teacher-generated outputs has proven effective for distilling various capabilities~\cite{chiang2023vicuna, taori2023alpaca, huang2022large, wang2022self, mitra2023orca}.
Beyond supervised approaches, reinforcement learning-based methods are also popular for knowledge distillation~\cite{bai2022constitutional, kim2023aligning, guo2024direct, luo2023wizardmath, agarwal2024policy, ko2024distillm}.
These methods often leverage preference data or ranking signals generated by teacher models.
Our work focuses on defending teachers and attacking students during output-based KD, where only input-output pairs are available to train the student model (i.e., ``black-box'' or API-access only). 
\fi

\noindent\textbf{Controllable Text Generation (CTG):}
CTG aims to steer LLM outputs to satisfy predefined conditions while maintaining fluency and coherence~\cite{liang2024controllable}.
Early approaches trained conditional LMs with explicit control codes that govern style and content~\cite{keskar2019ctrl}.
Inference-time methods such as gradient-based steering~\cite{madotto2020plug} and
discriminator-guided decoding~\cite{yang2021fudge,krause2021gedi} offer greater flexibility by modifying generation without retraining.
Prompt-based approaches are even more light-weight, employing 
%techniques such as 
chain-of-thought reasoning~\cite{wei2022chain}, directional stimulus~\cite{li2023guiding}, and iterative self-refinement~\cite{madaan2023self}.
%to enforce constraints on outputs without parameter updates. 

\smallskip
\noindent\textbf{Anti-Distillation:}
Recent work has explored proactive prevention of unauthorized distillation by manipulating model outputs.
Antidistillation Sampling (ADS)~\cite{savani2025antidistillation} is a sampling-based method that achieves a better trade-off between teacher utility and anti-distillation effectiveness compared to naive temperature sampling.
However, with sampling parameters effective for anti-distillation,
ADS often produces unnatural or incoherent text.
Defensive output generation (DOGe)~\cite{li2025doge} post-trains the teacher model's final layer to be inherently defensive against distillation.
While effective, DOGe's outputs are also sometimes unnatural.
Moreover, as a post-training approach, DOGe fundamentally lacks flexibility: the model is either defensive or not at all, and cannot adjust defense strength without retraining.
\citet{ding2025information} remove self-talk behaviors and reorder sub-conclusion ahead of the reasoning step, which is better at preserving semantics but has limited anti-distillation effects.
In contrast, our method requires no modification to the teacher model, guarantees semantic coherence in the generated traces, and achieves strong anti-distillation.

\smallskip
\noindent\textbf{Fingerprinting and Watermarking:}
Model fingerprinting aims to protect the model itself from unauthorized \emph{fine-tuning} (e.g., if the model is openly released but with a restrictive licensing agreement) by allowing model owners to uniquely identify their models~\cite{gu2022watermarking, xu2024instructional}.
On the other hand, 
model watermarking~\cite{liang2024watermarking,wan2022comprehensive} operates on model outputs.
Common \emph{text watermarking} approaches aim to determine whether the text was AI generated~\cite{kirchenbauer2023watermark,zhao2023provable}.
In contrast, \emph{API watermarking} methods 
are explicitly proposed as a defense against unauthorized knowledge distillation~\cite{he2022protecting,he2022cater,zhao2022distillation,zhao2023protecting}.
Many of the latter methods focus on traditional NLP tasks, such as sentiment analysis, and thereby assume that labels come from a simple structured space (e.g., real values or a small set of classes)~\cite{li2023plmmark,liu2023watermarking}. 
Furthermore, nearly all API watermarking approaches rely on token-level statistical techniques for detection which result in a non-trivial tradeoff between verification success and false alarm rates. 
And for those that do operate on sentence level~\cite{hou2024semstamp,dabiriaghdam2025simmark}, their transferability after distillation is unreliable, or they lack teacher-specific attribution necessary for proving unauthorized distillation.
In contrast, our proposed approach is both simpler and (as we show) substantially more reliable.

\section{Preliminaries}\label{sec:prelim}

\subsection{LLMs and Reasoning}\label{sec:prelim:subsec:llm_and_reasoning}

\noindent \textbf{Large Language Models (LLMs):} LLMs are neural networks trained on massive text corpora to predict the next token given provided context.
Formally, 
an LLM is a a parametric function $p_\theta$ with parameters $\theta$, mapping a sequence of input tokens $x_{1:t} = (x_1, x_2, \ldots, x_t)$, with $x_i$ from a vocabulary set $\mathcal{W}$, to a distribution over the next token.
Given any sequence of tokens as input,
the model computes the conditional probability distribution,
$p_\theta(\cdot | x_{1:t})$,
of all next-token probabilities.
%Thus a scalar probability of a particular next token $x_{t+1}$ is $p_\theta(x_{t+1} | x_{1:t})$.

%\smallskip
\noindent \textbf{Reasoning Traces:}
In this work, we define reasoning traces as structured outputs that explicitly decompose problem-solving processes into intermediate steps.
Formally, given a problem or query $q$, a reasoning trace (response) $r$ is a sequence $r = (s_1, s_2, \ldots, s_k, a)$,
where each $s_i$ represents an intermediate reasoning step, 
and $a$ is the final answer.
The generation of reasoning traces can be elicited through prompting techniques
(e.g., chain-of-thought prompting~\cite{wei2022chain}) or by training models explicitly to produce such structured outputs~\cite{guo2025deepseek, jaech2024openai}.

\subsection{Knowledge Distillation}\label{sec:prelim:subsec:kd}

Knowledge distillation (KD) is a technique for transferring knowledge from a large, capable teacher model to a smaller, more efficient student model~\cite{hinton2015distilling}.
In the context of LLMs, let $\mathcal{T}$ denote the teacher model and $\mathcal{S}$ denote the student model.
Given a dataset of queries $Q = \{q_1, q_2, \ldots, q_n\}$,
the goal is to train the student model $\mathcal{S}$ to emulate the teacher model $\mathcal{T}$'s behavior~\cite{xu2024survey}.
There are different training methods for knowledge distillation.
In this work, we focus primarily on supervised fine-tuning (SFT)-based distillation, as it is widely adopted in practice.
%\noindent \textbf{Supervised Fine-Tuning (SFT):}
In SFT-based distillation, the teacher is given a sequence of queries $Q = \{q_1, \ldots, q_n\}$, one at a time, and generates responses $r_i=\mathcal{T}(q_i)$ to each $q_i$.
These are then used to construct a dataset $D = \{(q_i, r_i)\}_{i=1}^n$.
The student model is then trained by minimizing 
\[
\mathcal{L}_{\text{SFT}}(\mathcal{S}; D) = -\sum_{i=1}^n \sum_{t=1}^{|r_i|} \log P_{\mathcal{S}}(r_i^{(t)} \mid q_i, r_i^{(<t)})
\]
where $r_i^{(t)}$ denotes the $t$-th token in trace $r_i$, and $r_i^{(<t)}$ denotes all preceding tokens.
%This objective encourages the student to learn both to produce correct final answers and to emulate the teacher's reasoning process.

\section{Model}\label{sec:model}

\subsection{Problem Setting}\label{sec:model:subsec:problem_formulation}

Consider an SFT-based knowledge distillation in which a student $\mathcal{S}$ sequentially submits $n$ queries $\{q_1, q_2, \ldots, q_n\}$ to the teacher $\mathcal{T}$, which responds with $r_i = \mathcal{T}(q_i)$.
This produces a dataset $D_{\text{clean}}=\{(q_i,r_i)\}_{i=1}^n$, which we refer to as ``clean'' to indicate that responses $r_i$ in this dataset are \emph{prior} to the rewriting techniques we discuss below.
Let $\mathcal{S}_{\text{clean}} = \text{Train}(D_{\text{clean}})$ denote the student trained on this data.
We suppose that the teacher is able to modify responses $r_i$ to alternative responses $r_i'$ using a \emph{rewriting method} $\mathcal{R}$ with $r_i' = \mathcal{R}(q_i,r_i)$.
This results in a modified dataset $D_{\mathcal{R}}=\{(q_i,r_i')\}_{i=1}^n$, which the developer then uses for training, obtaining a distilled student model $\mathcal{S}_{\mathcal{R}} = \text{Train}(D_{\mathcal{R}})$.
We define $\mathcal{T}_{\mathcal{R}}(q) \equiv \mathcal{R}(q,\mathcal{T}(q))$, that is, the teacher whose responses are rewritten by $\mathcal{R}$.
% \red{We define $\mathcal{T}_{\mathcal{R}}(q) \equiv \mathcal{R}(q,\mathcal{T}(q))$ as the rewriting-augmented teacher, which applies $\mathcal{R}$ to rewrite its original responses.}

%In typical sequence-level knowledge distillation for LLMs, 
%a teacher model provider controls a capable model $\mathcal{T}$, 
%and a student model developer seeks to train a smaller model $\mathcal{S}$ by learning from $\mathcal{T}$'s outputs,
%often on a dataset they possess: $Q=\{q_1, q_2, \ldots, q_n\}$.
%In standard instruction-following settings, each query $q_i$ is formatted as a combination of a system instruction $p_{\text{sys}}$ and a user query $q_i^{\text{user}}$. 
%The teacher model processes these inputs to generate responses: $r_i = \mathcal{T}(p_{\text{sys}}, q_i^{\text{user}})$.
%The student developer queries the teacher model with each problem and obtains corresponding reasoning traces, resulting in a dataset of problem-trace pairs $D_{\text{standard}}=\{(q_i,r_i)\}_{i=1}^n$.

%In this work, we focus on methods that teacher model providers can use to protect their model's intellectual property against unauthorized distillation. 
%At a high level, instead of directly outputting the clean trace $r_i$ for query $q_i$, the provider returns a modified trace $r_i'$ that is derived from the clean trace $r_i$ as the output. 

%We formulate the two approaches to protect model IP in more details next.

Our goal is to mitigate the risks associated with unauthorized LLM distillation.
We consider two means to this end: anti-distillation and API watermarking.
The former aims to rewrite the responses in order to degrade student training without compromising teacher accuracy.
The latter aims to embed an identifiable watermark in the generated response set.
A key constraints we impose in both cases is that the rewritten traces should preserve semantics of the original responses, which prevents modifications from being easily detectable.
We present formal problem statements for these next.

\subsubsection{Anti-Distillation}\label{sec:model:subsec:ad}

Anti-distillation aims to prevent unauthorized distillation by actively degrading the training efficacy of the traces, without significantly harming the teacher's performance.
This approach thereby discourages unauthorized distillation by making the resulting student model unreliable.
%The objective is to modify the traces such that student models trained on them exhibit significantly reduced performance.
Formally, let $\text{Acc}(\mathcal{S}, \mathcal{D})$ denote accuracy of a student model $\mathcal{S}$ on a target distribution $\mathcal{D}$ of query and answer pairs $(q,a)$.
Anti-distillation aims to design a rewriting procedure $\mathcal{R}$ to achieve
%then 
%The objective is to produce modified traces such that:
\begin{subequations}
\begin{align}
\text{Acc}(\mathcal{S}_{\text{clean}}, \mathcal{D}) - \text{Acc}(\mathcal{S}_{\mathcal{R}}, \mathcal{D}) &> \delta\label{C:1a}\\
\text{Acc}(\mathcal{T},\mathcal{D}) - \text{Acc}(\mathcal{T}_{\mathcal{R}},\mathcal{D}) &\le \epsilon.\label{C:1b}
\end{align}
\end{subequations}
for some large student performance degradation margin $\delta$ and small teacher performance degradation margin $\epsilon$ (where $\epsilon < 0$ indicates improved teacher performance).
\eqref{C:1a} provides for student performance degradation, while \eqref{C:1b} aims to limit teacher degradation.
In practice, we use an annotated test dataset $D$ as a proxy for a target distribution $\mathcal{D}$.
%Crucially, the teacher model itself must maintain its performance, ensuring that the modifications only affect the distillation process rather than the teacher's direct utility.

\begin{figure*}[ht]
    \centering
    \includegraphics[width=0.8\linewidth]{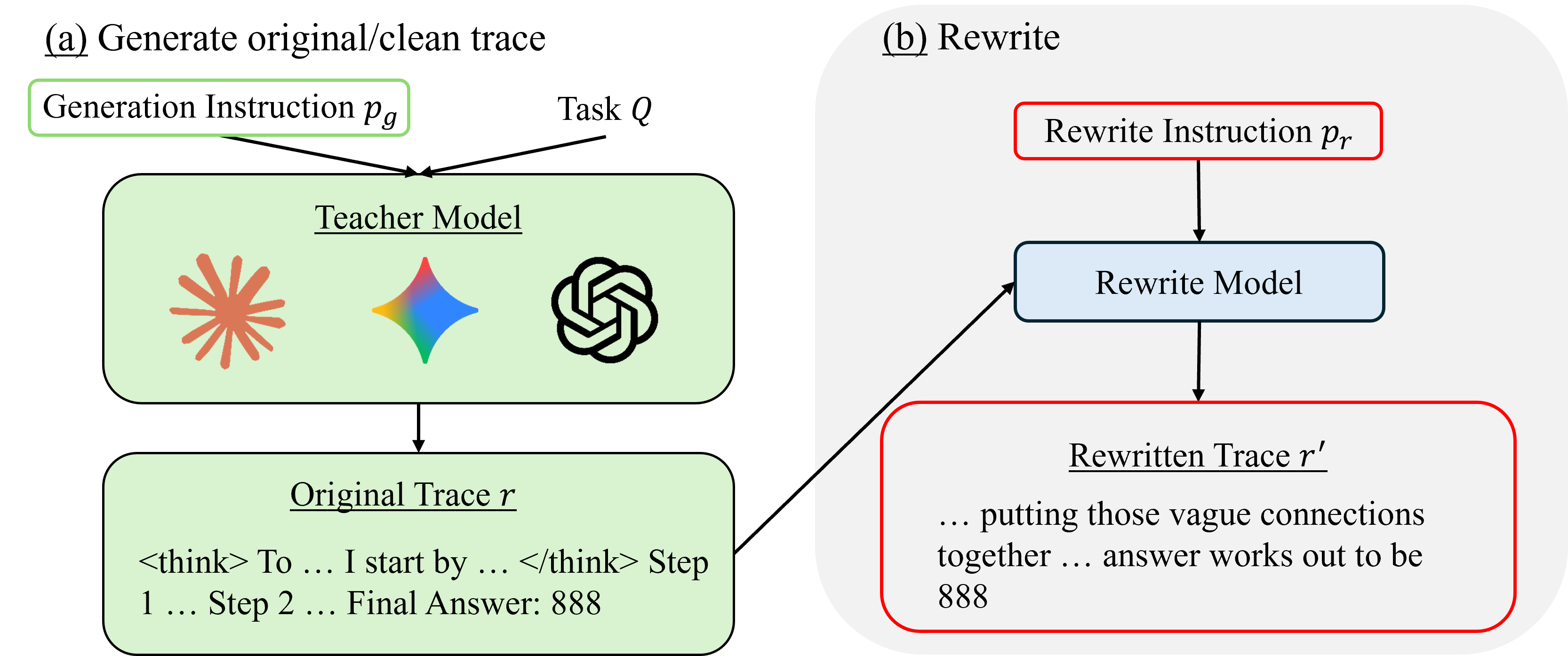}
    \caption{Overview of instruction-based rewriting.:
    (a) \textbf{Clean trace generation}: The teacher model $\mathcal{T}$ generates a reasoning trace $r$ for given task (query) $q$ using a standard generation instruction $p_g$. 
    (b) \textbf{Rewriting}: A rewrite model $\mathcal{R}$ with a rewrite instruction $p_r$ transforms $r$ into $r'$ to achieve IP protection while maintaining utility.}
    \label{fig:method_pipeline}
\end{figure*}

\subsubsection{API Watermarking}\label{sec:model:subsec:output_wm}

Output watermarking aims to embed verifiable signatures into the teacher model's reasoning traces such that student models trained on these traces inherit the detectable characteristics, and can be interactively verified to contain the watermark.
%This allows the teacher model provider to verify whether a suspect student model was trained using their outputs.

Formally, let $\mathcal{V}(\mathcal{S},\mu)$ denote a (possibly interactive) verification procedure that checks whether a student model $\mathcal{S}$ exhibits the watermark $\mu$ ($\mathcal{V}(\mathcal{S},\mu)=1$) or not ($\mathcal{V}(\mathcal{S},\mu)=0$).
Our objective of API watermarking  is to design a rewriting procedure $\mathcal{R}$ such that for a target watermark $\mu$,
\begin{subequations}
\begin{align}
&\Pr[\mathcal{V}(\mathcal{S}_{\mathcal{R}},\mu) = 1] \ge 1-\epsilon\label{C:2a}\\
&\Pr[\mathcal{V}(\mathcal{S}_{\text{clean}},\mu) = 1]  \le \epsilon\label{C:2b}\\
&\text{Acc}(\mathcal{S}_{\text{clean}}, \mathcal{D}) - \text{Acc}(\mathcal{S}_{\mathcal{R}}, \mathcal{D}) \le \epsilon\label{C:2c}\\
&\text{Acc}(\mathcal{T},\mathcal{D}) - \text{Acc}(\mathcal{T}_{\mathcal{R}},\mathcal{D}) \le \epsilon\label{C:2d}
\end{align}
\end{subequations}
for a small $\epsilon$.
\eqref{C:2a} ensures that the watermark is reliably detected; 
\eqref{C:2b} limits the false alarm rate; 
\eqref{C:2c} ensures that watermarking does not impact \emph{student accuracy}; and \eqref{C:2d} ensures that it does not impact \emph{teacher accuracy}.
%where $\mathcal{S}_{\text{clean}}$ denotes a student model trained on unmodified traces.
%The first constraint ensures the watermark is reliably detectable in models trained on the modified traces, 
%while the second constraint ensures that the watermark does not significantly degrade the student model's utility on standard tasks.

\subsection{Constraints on Rewriting}\label{sec:model:subsec:constraints}

We impose two constraints on rewriting $\mathcal{R}$ that reflect realistic deployment scenarios.

\smallskip
\noindent \textbf{Limited Control Scope:}
The teacher model providers have control only over the reasoning traces generated by their model. 
They have no influence over the student training process, including the choice of student model architecture and hyperparameters.
Furthermore, the teacher cannot modify the sequence of queries $Q$, insert additional training examples, or alter the dataset composition in any way other than transforming its own generated traces.
Accordingly, we focus on single-source distillation, where the distiller wants to replicate the capabilities of a specific teacher model---the scenario where unauthorized distillation is most practically relevant. For example, we wish to prevent a malicious actor from distilling frontier close-sourced models such as ChatGPT or Claude.

\smallskip
\noindent \textbf{Trace Quality Preservation:}
We require the response modifications to preserve \emph{both} the correctness (of the answer $a$) \emph{and} the semantic quality of the full response $r$.
%performance of the teacher model,  echoing the ``clean-label'' constraint widely studied in adversarial machine learning~\cite{vorobeychik2018adversarial}.
%While ``clean-label'' generally refers to adversarial samples in classification tasks that retain correct labels, we adapt this concept to the reasoning trace setting.
%In our context, clean-label requires not only answer correctness but also semantic coherence: modified traces must remain well-formed and logically plausible.
This constraint rules out trivial strategies such as injecting random tokens or nonsensical phrases, ensuring that modified traces can pass reasonable quality controls while still achieving their objective.

\section{Methodology}\label{sec:method}

% In this section, 
% we present two classes of approaches that operate at different levels of abstraction and control.
% The first is \emph{instruction-based rewriting}, which uses an LLM assistant to rewrite the original responses, and can be applied to both anti-distillation and watermarking objectives.
% The second is \emph{gradient-based rewriting}, in which we use gradient methods to explicitly optimize the target objective, and which we use solely for anti-distillation.
% %to achieve  objectives while respecting the constraints.
% We describe each in further detail below.

In this section, 
we present two classes of approaches that operate at different levels of abstraction and control.
The first is \emph{instruction-based rewriting}, which uses an LLM assistant to rewrite the original responses, and can be applied to both anti-distillation and watermarking objectives.
The second is \emph{gradient-based rewriting}, in which we use gradient methods to explicitly optimize the target objective, and which we use solely for anti-distillation.
Conceptually, gradient-based rewriting serves as the principled starting point that directly optimizes for student degradation, while instruction-based rewriting offers a more practical alternative that leverages LLMs' semantic understanding of reasoning quality to achieve desired trace manipulation.
We describe each in further detail below.

\subsection{Instruction-Based Rewriting}\label{sec:method:subsec:instruction_rewriting}

Instruction-based rewriting implements $\mathcal{R}(q,r)$ by querying an assistant LLM,
%realizes the modification to the clean traces through a rewrite LLM with natural language instructions.
as illustrated in Figure~\ref{fig:method_pipeline}.
We explore two methods for designing effective rewrite instructions: semantic prompting and optimized prompting.

\subsubsection{Semantic Prompting}
The simplest approach uses a natural language prompt $p$ that describes the desired transformation at a high semantic level. 
Despite the simplicity, we find that semantic prompting is remarkably effective when executed by capable language models.
The key insight is that modern LLMs appear to possess sufficient understanding of reasoning quality to be able to implicitly degrade it through high-level directives.

\smallskip
\noindent \textbf{Application to Anti-Distillation:} For anti-distillation, the prompt we use directly specifies the objective to an LLM rewriting assistant $\mathcal{A}$ (see the Supplement for the full prompt).
Notably, $\mathcal{R}$ in this case depends only on the trace $r$ to be rewritten, i.e., $\mathcal{R}(q,r) = \mathcal{A}(p_r,r)$, where $p_r$ is the rewrite instruction to $\mathcal{A}$.

\smallskip
\noindent \textbf{Application to API Watermarking:} For API watermarking, the rewrite instruction to the LLM assistant $\mathcal{A}$ contains a target watermark $\mu$ that, in this work, takes the form ``\texttt{trigger} = \texttt{target}''.
Thus, in this case $\mathcal{R}(q,r) = \mathcal{A}(p_r,r)$.
In principle, both the trigger and the target can be arbitrary.
In the Supplement, we explore the relative efficacy of several strategies for generating these.

\subsubsection{Optimized Prompting}

Recent work has shown that LLMs can effectively and automatically optimize prompts~\cite{guo2023connecting,wang2023promptagent,yang2023large,zhou2022large}.
%Since our rewrite instructions are themselves prompts to the rewrite model,
%a natural question is whether automated prompt optimization can discover more effective instructions.
Consequently, the next step from direct semantic prompting is to design optimized prompts for our task.
We do this by adapting the Optimization by PROmpting (OPRO) framework~\cite{yang2023large}.
%,
%which formulates prompt optimization as iterative meta-learning.
Specifically, in each step $k$, we maintain a history of prompt-score pairs $H_k = \{(p^{(i)}, s^{(i)})\}$,
where $s^{(i)}$ measures the effectiveness of prompt $p^{(i)}$.
An optimizer LLM 
%$\mathcal{O}$ 
uses this history to propose $m$ new candidate instructions $\{p^{(k,1)}, \ldots, p^{(k,m)}\} = \mathcal{O}(H_k)$.
%\[
%\{p^{(k,1)}, \ldots, p^{(k,m)}\} = \mathcal{O}(H_k)
%\]
Each candidate prompt $p^{(k,j)}$ is then evaluated using a scoring function $f(p)$, which quantifies its success.
In particular, we define the following score function for anti-distillation, which makes use of a set of proxy student models $\mathbf{S}_{\text{proxy}}$ (and is normalized by its cardinality):
\begin{align*}
    f(p) = \sum_{\mathcal{S} \in \mathbf{S}_{\text{proxy}}} [\text{Acc}(&\mathcal{S}_{\text{clean}}, \mathcal{D}) - 
    \text{Acc}(\mathcal{S}_{\mathcal{R}_p}, \mathcal{D})]
\end{align*}
where $\mathcal{S}_{\mathcal{R}_p}$ refers to a proxy student model $\mathcal{S}$ trained on the data with responses $r$ rewritten by $\mathcal{R}_p = \mathcal{A}(p,r)$.
As before, since we don't have access to the target distribution $\mathcal{D}$, we use a validation dataset to approximate $f(p)$.
%$\mathcal{M}_{\text{proxy}}$ is an ensemble of proxy student models (distinct from the actual victim model),
%$M_{\text{clean}}$ denotes a model trained on clean traces,
%$M_{p_r}$ denotes a model trained on traces rewritten with instruction $p_r$,
%and $D_{\text{val}}$ is a validation set.

\begin{figure*}[t]
    \centering
    \includegraphics[width=0.95\linewidth]{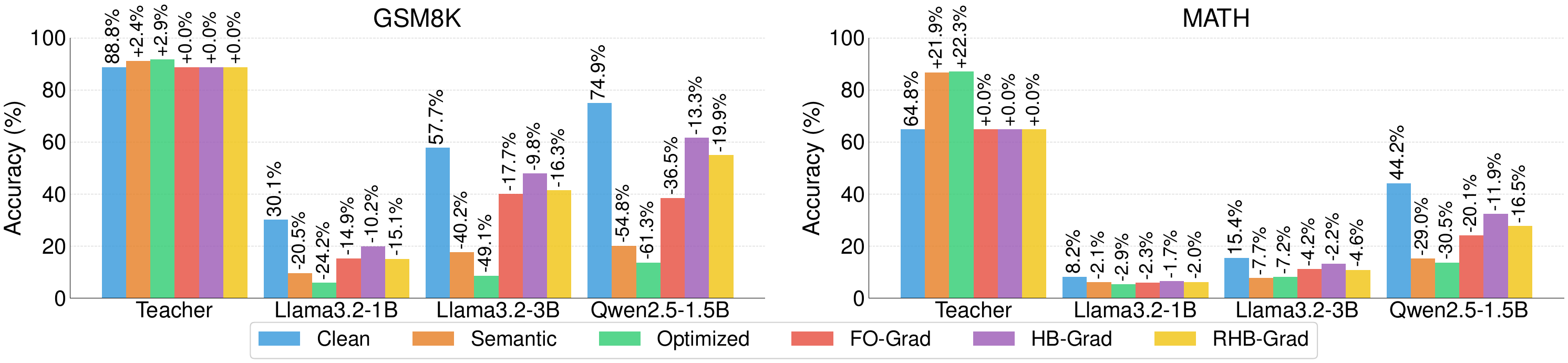}
    \caption{Comparison of our rewriting approaches for anti-distillation on GSM8K (left) and MATH (right).}
    \label{fig:grad_embed_comparison}
\end{figure*}

\iffalse
\begin{figure*}[t]
    \centering
    \includegraphics[width=0.95\linewidth]{assets/approach_comparison.pdf}
    \caption{Comparison of our rewriting approaches for anti-distillation on GSM8K (left) and MATH (right).}
    \label{fig:our_approach_comparison}
\end{figure*}
\fi

\begin{figure*}[t]
    \centering
    \includegraphics[width=0.85\linewidth]{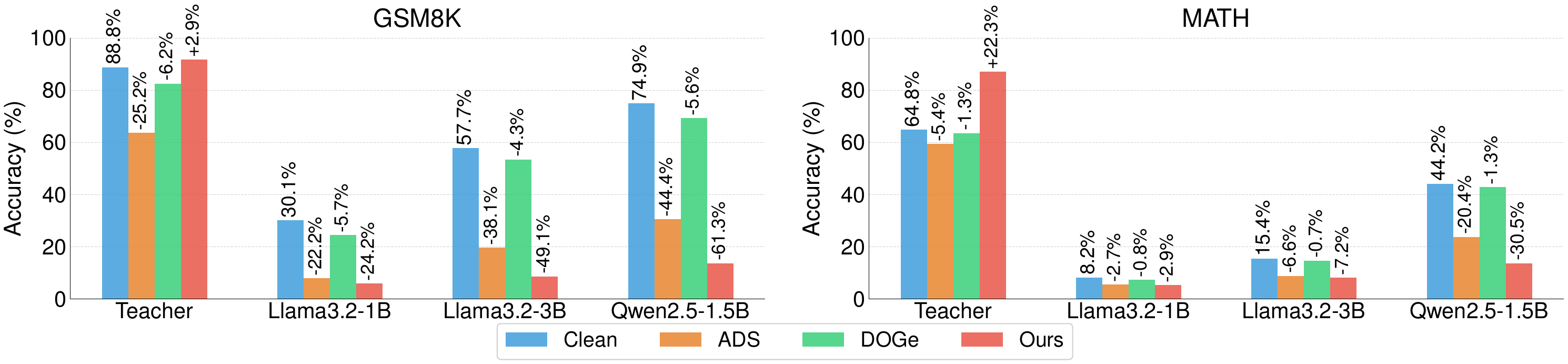}
    \caption{Anti-distillation comparisons on GSM8K (left) and MATH (right). 
    Our method achieves the strongest anti-distillation effect without compromising the teacher's utility.}
    \label{fig:ad}
\end{figure*}

\subsection{Gradient-Based Rewriting}\label{sec:method:subsec:gradient_rewriting}

In addition to LLM-assisted rewriting above, we also develop gradient-based rewriting methods, 
which can in principle provide finer-grained control over trace manipulation by directly optimizing for the objective.
On the other hand, since we do not know the actual student a priori and must use a proxy student $\mathcal{S}_{\text{proxy}}$ (or a collection thereof) in its place, there is a risk that such approaches may also overfit to the proxy students.
%These approaches require access to a differentiable proxy student model, $\mathcal{S}_{\text{proxy}}$.

\iffalse
Formally, we are
given a clean trace $r = (r^{(1)}, r^{(2)}, \ldots, r^{(T)})$ where $r^{(t)}$ denotes the $t$th token.
%the objective is to find a trace $r'$ that maximizes the distillation loss that a student model would incur when training on $r'$.
For an anti-distillation objective, we seek to construct a rewritten trace $r'$ that maximizes
\(
\mathcal{L}(r'; q) = \mathcal{L}_{lm}(\mathcal{S}_{\text{proxy}}, (q, r'))
\)
subject to the constraints defined in Section~\ref{sec:model:subsec:constraints}, where $\mathcal{L}_{\text{lm}}$ refers to standard cross-entropy loss.
Henceforth, we omit the $\text{lm}$ subscript.
We consider two approaches detailed next for solving this problem.
\fi

\subsubsection{Embedding-Space Poisoning}

Our first gradient-based approach modifies the embedding of tokens, taking inspiration from gradient-based poisoning attacks~\cite{vorobeychik2018adversarial}.
However, typical attacks of this kind make use of the implicit function theory to approximate gradients with respect to data (in our case, trace embedding) modifications, which requires computing an inverse of the loss Hessian; these are infeasible at scale, such as for LLMs.
To address this, we propose an approximation that eliminates the need for the inverse Hessian computation.
%using the Projected Gradient Descent (PGD) framework~\cite{vorobeychik2018adversarial}.
%The key insight is that data poisoning is fundamentally a bilevel optimization problem: we seek trace modifications that, when the student \emph{trains} on them, result in poor performance on held-out test data.

Specifically, consider a trace $r$ and let $\mathbf{E} = (\mathbf{e}^{(1)}, \ldots, \mathbf{e}^{(T)})$ represent its embedding sequence, where $\mathbf{e}^{(t)} \in \mathbb{R}^d$ is the embedding of token $r^{(t)}$.
Let $\theta$ denote a (proxy) student's parameters and $\eta$ the learning rate.
Our objective is to maximize the test loss:
%under these updated parameters:
\[
\max_{\mathbf{E}'} \mathcal{L}_{\text{test}}(\mathbf{E}') \equiv \mathcal{L}(\theta(\mathbf{E}'); D_{\text{test}})
\]
where $D_{\text{test}}$ is a held-out set of examples and $\theta(\mathbf{E}')$ are the parameters of the student's model after fine-tuning with a modified trace $\mathbf{E}'$.
The gradient of this objective is
%To compute gradients of this objective with respect to $\mathbf{E}'$, we differentiate with respect to $\mathbf{E}'$:
%through the parameter update step:
\[
\nabla_{\mathbf{E}'} \mathcal{L}(\theta(\mathbf{E}'); D_{\text{test}}) = \nabla_{\theta} \mathcal{L}(\theta_0) \cdot \frac{d \theta(\mathbf{E}')}{d \mathbf{E}'},
\]
where $\theta_0$ are the pre-trained student parameters.
The main issue is approximating $\frac{d \theta(\mathbf{E}')}{d \mathbf{E}'}$.
Suppose that we take a single gradient descent iteration on a modified trace with embeddings $\mathbf{E}'$:
%, the updated parameters are:
\(
\theta' = \theta - \eta \nabla_{\theta} \mathcal{L}(\theta; \mathbf{E}').
\)
We can then approximate
\(
\frac{d \theta(\mathbf{E}')}{d \mathbf{E}'} \approx - \eta \nabla^2_{\theta, \mathbf{E}'} \mathcal{L}(\theta_0; \mathbf{E}'),
\)
where
%where the term $\frac{\partial \theta'}{\partial \mathbf{E}'}$ involves the mixed Hessian 
$\nabla^2_{\theta, \mathbf{E}'} \mathcal{L}$ is the mixed Hessian of the loss.
Then, we iteratively update the trace embeddings as 
%follows:
\[
\mathbf{E}^{(k+1)} = \Pi_{\epsilon}\left(\mathbf{E}^{(k)} + \alpha \cdot \text{sign}\left(\nabla_{\mathbf{E}^{(k)}} \mathcal{L}_{\text{test}}%(\mathbf{E}^{(k)})
\right)\right)
\]
where
$\alpha$ is the step size and $\Pi_{\epsilon}(\cdot)$ projects the perturbed embeddings back into an $\ell_\infty$ ball of radius $\epsilon$ around the original embeddings $\mathbf{E}$.

An important limitation of this approach is that it is still computationally expensive as it requires Hessian computation.
Note, however, that our goal is to make the traces \emph{difficult} to train from, and this is a property we can often expect from \emph{adversarial input perturbations}~\citep{tran2018spectral}.
This leads to an alternative iterative update scheme with $\nabla_{\mathbf{E}^{(k)}} \mathcal{L}_{\text{test}}$ replaced with $\nabla_{\mathbf{E}^{(k)}} \mathcal{L}$,
%\[
% \mathbf{E}^{(k+1)} = \Pi_{\epsilon}(\mathbf{E}^{(k)} + \alpha \cdot \text{sign}(\nabla_{\mathbf{E}^{(k)}} \mathcal{L}))
%\]
where $\mathcal{L}$  is the cross-entropy loss of the (proxy) student model.
In effect, we can view this as the following approximation of the objective above: $\mathcal{L}_{\text{test}}(\mathbf{E}') \approx \mathcal{L}(\theta_0,\mathbf{E}')$.
We refer to the former approach as \emph{Hessian-based (HB-Grad)} and the latter as \emph{first-order (FO-Grad)} gradient-based rewriting.
Additionally, we consider a robust variant (\emph{RHB-Grad}) of HB-Grad that adds Gaussian noise to the proxy student's parameters before computing the gradient.

After $K$ iterations, we project the final perturbed embeddings back to the discrete token space.
For each perturbed embedding $\mathbf{e}'^{(t)}$, we select the token whose embedding is nearest:
\[
r'^{(t)} = \underset{v \in \mathcal{V}}{\arg\min} \|\mathbf{e}'^{(t)} - \text{Embed}(v)\|_2
\]
where $\mathcal{W}$ is the vocabulary and $\text{Embed}(\cdot)$ is the embedding function.

\subsubsection{Satisfying Constraints}
%Both gradient-based approaches face the challenge of satisfying our constraints.
%while optimizing the attack objective.
To preserve the correctness of rewriting, we mask the final answer in the trace during gradient-based optimization, so that it is not modified by gradient updates.
%The HotFlip approach naturally preserves semantic coherence if the number of flips is small.
%As long as we constrain the number of flips to be small, the overall coherence is maintained.
We additionally constrain $\alpha$ and $\epsilon$ to be small to limit the semantic impact.
%of rewriting.
%large semantic deviations from the original trace.
\section{Experiments}\label{sec:experiments}

This section begins with descriptions of our experimental setups,
we then present our results organized into two parts where the first addresses anti-distillation output generation (Section~\ref{sec:exp:subsec:ad}), 
and the second output watermarking for IP protection (Section~\ref{sec:exp:subsec:wm}).

\subsection{Setup}

\noindent \textbf{Models:}
We use
\texttt{DeepSeek-R1-Distill-Qwen-\\7B} as the teacher model and
\texttt{gpt-oss-120b} as the rewrite model.
In anti-distillation, 
\texttt{Llama-3.2-3B},
\texttt{Llama-3.2-1B}, 
and \texttt{Qwen2.5-1.5B} are used as the student models.
In API watermarking,
\texttt{Llama-3.2-3B},
\texttt{Llama-3.1-8B},
and \texttt{Qwen2.5-1.5B} are used as the student models.
Full implementation details are in the Supplement.
%Appendix~\ref{appendix:implementation}.

\smallskip
\noindent \textbf{Datasets:}
To verify the effectiveness of our approach,
we evaluate it on four datasets:
GSM8K~\cite{cobbe2021training}
(using GSM8K Platinum~\cite{vendrow2025large} as test set) and
MATH~\cite{hendrycksmath2021} in the main paper, with the results on
MMLU~\cite{hendrycks2020measuring}
and MMLU-Pro~\cite{wang2024mmlu} provided in the Supplement.
%\red{main body no longer has 4 dataset experiments, only GSM8K and MATH now}

\smallskip
\noindent \textbf{Evaluation Metrics:}
Our primary metric is zero-shot answer accuracy.
We aim to maximize it for teacher and minimize for student models in anti-distillation, while maximizing for both in watermarking.
To ensure consistent answer extraction across all models and datasets,
we adopt the answer forcing technique following~\citet{savani2025antidistillation} (see the Supplement for further details).
We measure efficacy of watermarking using true detection (TD) and false alarm (FA) rates.
The former measures the fraction of attempts in which the watermark is detected successfully for a distilled model, while the latter measures the same quantity for an undistilled model.

%For evaluating effectiveness, we compare performance between student models distilled from clean traces and our rewritten traces to demonstrate \textit{anti-distillation}.
%For \textit{watermarkings}, we verify if distilled students exhibit target behaviors in their outputs.

\begin{table*}[h]
\centering
\caption{Watermark detection results  on GSM8K. Teacher column shows teacher accuracy. For each student model, we report true detection rate (TD, left) and false alarm rate (FA, right). Each cell contains two values corresponding to $K=5$ and $K=20$ test queries, respectively. Bold indicates that the result is within 0.02 of the best.}
\label{tab:wm}
\resizebox{0.95\linewidth}{!}{%
\begin{tabular}{@{}lc|cccccc@{}}
\toprule
Method & Teacher & \multicolumn{2}{c}{Llama-3.1-8B} & \multicolumn{2}{c}{Llama-3.2-3B} & \multicolumn{2}{c}{Qwen2.5-1.5B} \\
\cmidrule(lr){1-2} \cmidrule(lr){3-4} \cmidrule(lr){5-6} \cmidrule(lr){7-8}
\textit{Clean} & 88.76\% & TD ($\uparrow$) & FA ($\downarrow$) & TD ($\uparrow$) & FA ($\downarrow$) & TD ($\uparrow$) & FA ($\downarrow$) \\ 
\midrule
He et al.~\cite{he2022protecting}    & 88.76\% & 0.94 / \textbf{1.00} & 0.76 / 1.00 & \textbf{0.95} / \textbf{1.00} & 0.80 / 1.00 & \textbf{0.99} / \textbf{1.00} & 0.78 / 0.99 \\
GINSEW~\cite{zhao2023protecting}     & 69.03\% & 0.01 / 0.06 & \textbf{0.02} / 0.16 & 0.01 / 0.04 & \textbf{0.01} / 0.13 & 0.01 / 0.08 & \textbf{0.01} / 0.07 \\
KGW~\cite{kirchenbauer2023watermark} & 71.15\% & 0.00 / 0.00 & \textbf{0.00} / \textbf{0.00} & 0.00 / 0.00 & \textbf{0.00} / \textbf{0.01} & 0.39 / 0.88 & \textbf{0.00} / \textbf{0.00} \\
VIA~\cite{liang2025virus}            & 88.76\% & 0.77 / \textbf{1.00} & \textbf{0.00} / \textbf{0.02} & 0.27 / 0.52 & \textbf{0.00} / \textbf{0.00} & 0.84 / \textbf{1.00} & \textbf{0.00} / \textbf{0.00} \\
\midrule
Ours & \textbf{90.98}\% & \textbf{1.00} / \textbf{1.00} & \textbf{0.00} / \textbf{0.02} & 0.55 / \textbf{0.99} & \textbf{0.00} / \textbf{0.00} & \textbf{0.98} / \textbf{1.00} & \textbf{0.00} / \textbf{0.00} \\
\bottomrule
\end{tabular}
}
\end{table*}
\subsection{Anti-Distillation Results}\label{sec:exp:subsec:ad}

In Figure~\ref{fig:grad_embed_comparison}, we evaluate the relative efficacy of the three proposed anti-distillation rewriting methods: two that are prompt-based (semantic and optimized), and one gradient-based which has three variants.
Here, we note two key findings.
First, \emph{all methods substantially reduce distillation efficacy while maintaining accuracy of the teacher model}.
Second, \emph{prompt-based approaches significantly outperform gradient-based rewriting}.
Notably, the best rewriting method is \emph{optimized prompting (OPT)}, which yields student accuracy below 20\% on both GSM8K and MATH datasets, with as much as $\sim$61\% accuracy reduction compared to distillation from clean traces.
Moreover, it actually \emph{increases} the accuracy of the teacher on GSM8K (by $\sim$3\%) and MATH (by $\sim$22\%) datasets.
This improvement is due to rewriter LLM being able to correct errors in the teacher's original traces during rewriting.

%We first compare the effectiveness of our proposed approaches.
%Figure~\ref{fig:grad_embed_comparison} compares the anti-distillation effectiveness of our proposed approaches.
%Across both datasets, all rewriting approaches successfully preserve teacher accuracy---our instruction-based rewriting approaches improve teacher accuracies (especially on MATH dataset), while substantially degrading student performance.
%We observe that the instruction-based rewriting with optimized instruction achieves the most effective attack performance.
%\emph{All subsequent reported results use this method unless otherwise specified.}

Next, Figure~\ref{fig:ad} compares \emph{OPT} (our best approach) to two recent anti-distillation baselines:  
%Next, we compare with two baselines:
(1) Antidistillation Sampling (ADS)~\cite{savani2025antidistillation}, 
%which poisons the teacher model's outputs by modifying the next-token probability distribution during inference;
and (2) DOGe~\cite{li2025doge}.
%, which shares similar adversarial insights with ADS but embeds the antidistillation capability directly into the teacher model by fine-tuning its final linear layer.
%Figure~\ref{fig:ad} shows the anti-distillation results comparing our method against the baselines.
We note two key findings: 1) \emph{OPT} yields \emph{considerably higher teacher accuracy}, and 2) \emph{OPT} has a \emph{consistently stronger anti-distillation effect}, compared to the baselines.
%We observe that, first, our method is the only one that preserves---and even improves---the teacher model's accuracy on both datasets.
%Second, across all three student models, our method achieves the most effective anti-distillation, producing the largest degradation in student model performance.
In addition, we find that \emph{OPT} maintains a strong anti-distillation effect as we use more capable students and with adaptive distillation (see Section~\ref{sec:exp:subsec:robust} and Appendix~\ref{appendix:ad_exp}).

\subsection{API Watermarking Results}\label{sec:exp:subsec:wm}

We compare our instruction-based rewriting approach with four state-of-the-art API watermarking baselines:
(1) \citet{he2022protecting}, which uses synonym replacements to the original outputs;
(2) GINSEW~\cite{zhao2023protecting}, which injects a secret sinusoidal signal into the model’s generation probabilities;
(3) KGW~\cite{kirchenbauer2023watermark}, which adds a bias to a pre-selected set of tokens;
and (4) Virus Infection Attack (VIA)~\cite{liang2025virus}, which directly injects target messages into text.
For (1)-(3), the teacher employs watermarking on all distillation.
For (4) and our method, we inject the watermark message into $10\%$ of the traces.

\begin{figure}[h]
    \centering
    \includegraphics[width=0.99\linewidth]{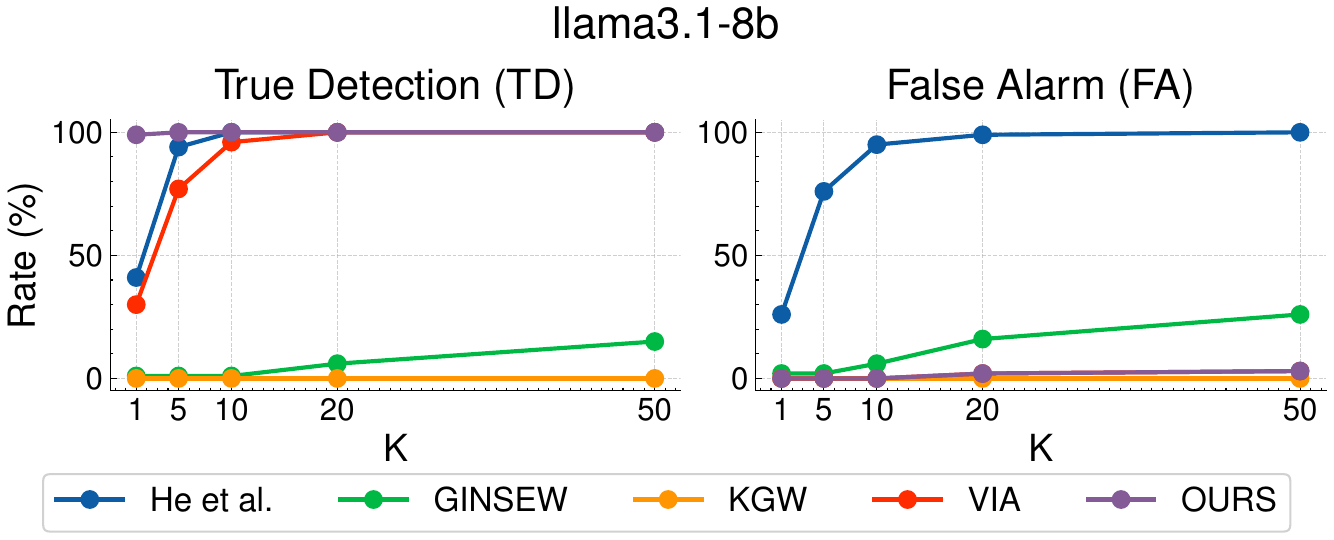}
    \caption{Watermark detection: true detection rate and false alarm rates vs. K for llama3.1-8B suspect student model.}
    \label{fig:main-wm-fig}
\end{figure}

For each value of $K$, we run $100$ independent trials with randomly sampled prompts and report the empirical TD and FA rates in Table~\ref{tab:wm},
% Table~\ref{tab:wm} presents the results, 
demonstrating that \emph{our approach nearly always yields the best or near-best performance in (a) teacher accuracy, (b) true detection, and (c) false alarm rates}.
Indeed, we are able to achieve near-perfect verification rate with very few ($K=5$) client queries, with \emph{zero false-alarm rate} with the exception of the least capable student (Llama-3.2-3B).
While He et al.~exhibits a high TD, its FA rate is unacceptably high.
VIA is, on balance, the most competitive baseline, but our approach is considerably more sample efficient, as we can observe in Figure~\ref{fig:main-wm-fig}, where we can achieve nearly perfect detection rate with no false alarms for only $K=1$ queries, while VIA's verification rate remains $\sim$30\%.

%To evaluate watermark detectability,  we generalize detections for baselines (1)-(3) from their respective token level statistical thresholds to per sample binary decision. 
%For (4) and our method, we directly verify if \texttt{target} is present in the model output. 
%Detection success is defined as at least one sample containing the watermark among $K$ test prompts.
%True detection (TD) refers to successful detection in students distilled from the watermarked dataset,
%while false alarm (FA) refers to detection in students distilled from the clean dataset.
%For each value of $K$, we run $100$ independent trials with randomly sampled prompts and report the empirical TD and FA rates in Table~\ref{tab:wm}.
%First, our method is best at preserving teacher performance. 
%More importantly, it is the only approach that allows reliable reliable true detection with negligible false alarms. 
%In contrast, while He et al.'s method has better one-shot TD on \texttt{Llama3.2-3B} model, its utility is compromised by excessive false positive rates, making the detection unreliable.
%\red{Exact implementations of baselines and their respective detection algorithms are documented in Appendix, along with results with larger $K$ values.}

\subsection{Robustness to Adaptive Distillation}\label{sec:exp:subsec:robust}

\begin{figure*}[t]
    \centering
    \includegraphics[width=0.9\linewidth]{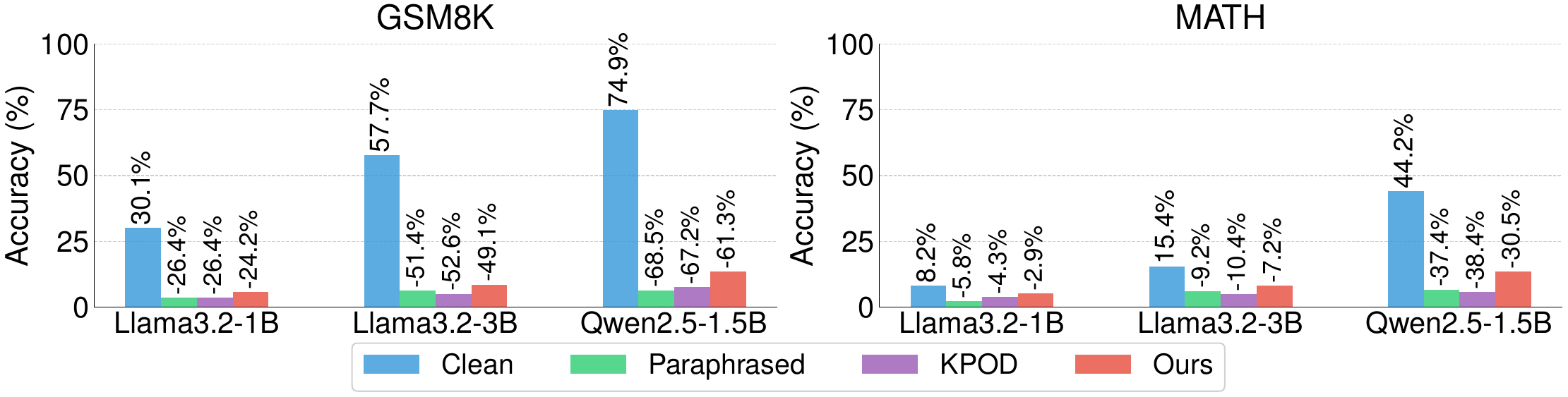}
    \caption{Robustness of anti-distillation to adaptive attacks. \textbf{Paraphrased}: distiller paraphrases our \emph{OPT} traces before fine-tuning. \textbf{KPOD}: distiller applies keypoint-based progressive distillation on our \emph{OPT} traces. \textbf{Ours}: standard SFT on our \emph{OPT} traces.}
    \label{fig:ad_robust}
\end{figure*}

\begin{figure}[t]
    \centering
    \includegraphics[width=0.99\linewidth]{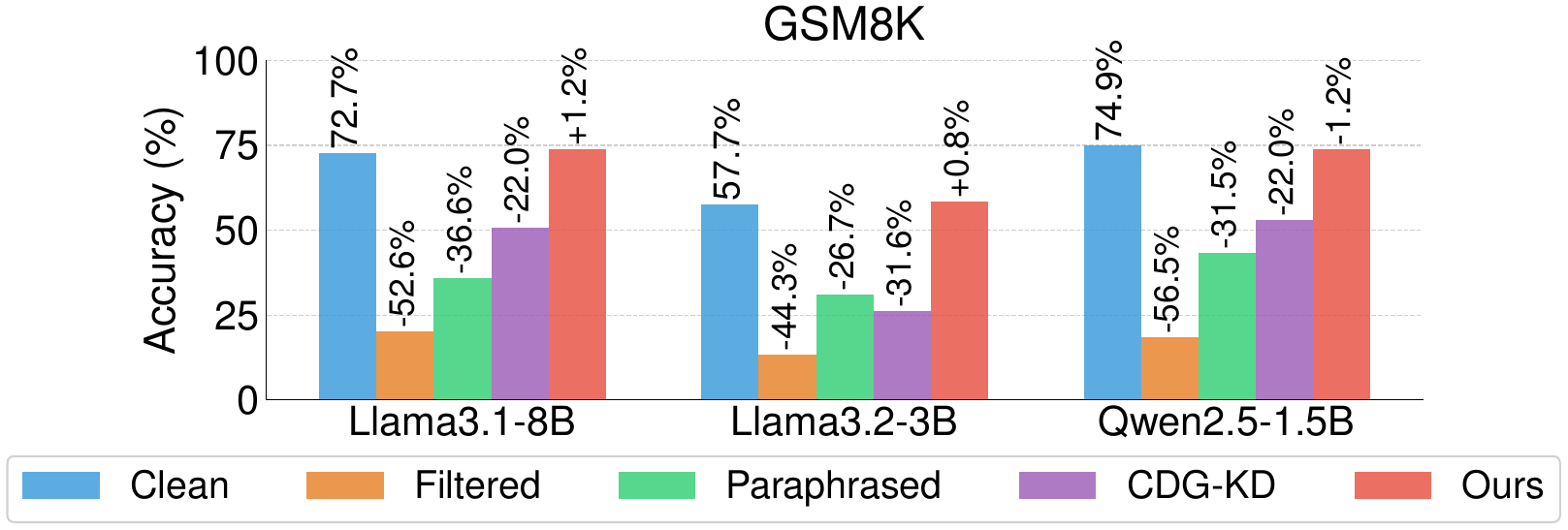}
    \caption{Student accuracy on GSM8K after distillation under adaptive attacks. \textbf{Clean}: distilled from original traces without watermarks. \textbf{Filtered}: distilled from watermark-injected traces after regex-based filtering that removes $\pm 3$ tokens around each \texttt{=} sign. \textbf{Paraphrased}: distilled from watermark-injected traces paraphrased by Parrot paraphraser. \textbf{CDG-KD}: distilled from traces processed by CDG-KD. \textbf{Ours}: distilled from watermark-injected traces.}
    \label{fig:wm_robust_student_perf}
\end{figure}

\begin{figure}[t]
    \centering
    \includegraphics[width=0.99\linewidth]{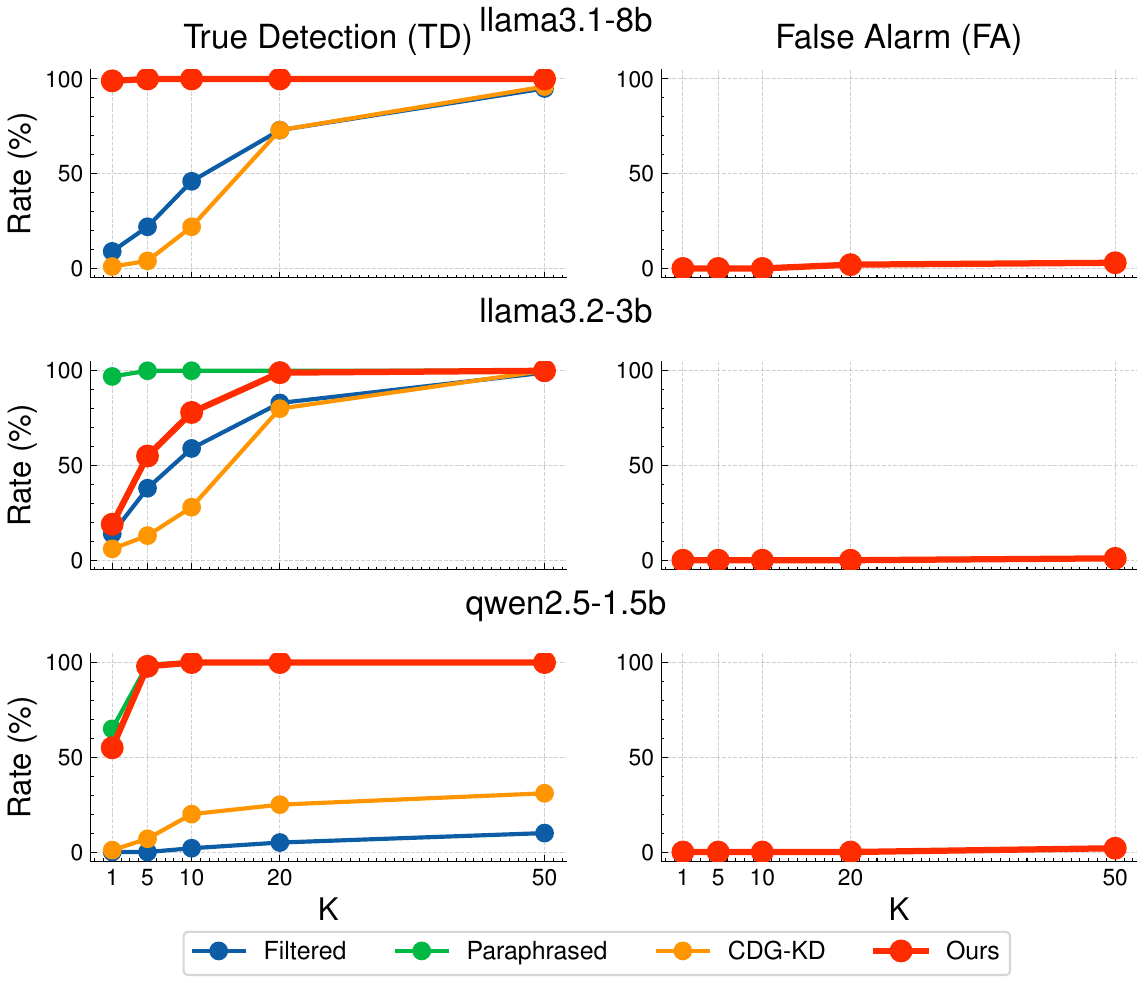}
    \caption{Watermark detection results under adaptive attacks (Filtered, Paraphrased, CDG-KD). The watermark remains detectable under all three attacks, while they substantially degrade student task accuracy (see Figure~\ref{fig:wm_robust_student_perf}).}
    \label{fig:wm_robust_detection}
\end{figure}

We now investigate whether our rewriting approaches for both anti-distillation and watermarking  remain effective when a distiller employs adaptive distillation strategies.
We make two threat model assumptions: the distiller targets a single source model, and has no prior knowledge of the specific watermark text, which is reasonable given that watermarks are injected into only 10\% of traces, and can be chosen arbitrarily.

\smallskip
\noindent 
\textbf{Anti-Distillation.}
We consider two adaptive attacks.
First, \textbf{Paraphrased}: the distiller paraphrases collected traces before fine-tuning using the Parrot paraphraser~\cite{prithivida2021parrot}.
As shown in Figure~\ref{fig:ad_robust}, paraphrasing not only fails to recover distillation efficacy but actually amplifies the anti-distillation effect, since paraphrasing tends to further destroy the structured format of reasoning traces.
Second, \textbf{KPOD}~\cite{feng2024kpod}: the distiller applies keypoint-based progressive chain-of-thought distillation, which tries to upweight more informative steps in a reasoning trace during training.
As shown in Figure~\ref{fig:ad_robust}, KPOD does not mitigate the anti-distillation effect; in fact, when applied to our rewritten traces, upweighting also amplifies the degradation relative to standard SFT.

\smallskip
\noindent 
\textbf{Watermarking.}
We consider three adaptive attacks: \textbf{Filtered}, where the distiller applies regex-based filtering that removes the 3 tokens surrounding every \texttt{=} sign in each trace; \textbf{Paraphrased}, where the distiller paraphrases traces before fine-tuning using the Parrot paraphraser~\cite{prithivida2021parrot}; and \textbf{CDG-KD}~\cite{yi2025cdgkd}, a contrastive decoding-guided distillation framework designed to scrub token-level statistical watermarks.
As shown in Figure~\ref{fig:wm_robust_student_perf}, all three attacks degrade student task performance, with filtering causing the most severe degradation.
More complicated strategies like CDG-KD also incur substantial accuracy drops since it effectively replaces teacher traces with weaker student-generated outputs.
Moreover, as shown in Figure~\ref{fig:wm_robust_detection}, the watermark remains detectable under all three attacks.
Notably, the paraphrase attack actually \emph{strengthens} detectability, likely because paraphrasing reinforces rather than removes the embedded semantic association.
CDG-KD fails to scrub our watermark as well because its mechanism targets token-level distributional shifts --- the signature of statistical watermarks --- whereas our watermark is a behavioral trigger that produces no such shift on normal inputs and activates only on a secret trigger.
Taken together, these results highlight a fundamental trade-off: any intervention aggressive enough to meaningfully reduce watermark detectability also destroys the reasoning quality that motivated distillation in the first place.
\section{Conclusion}

We proposed a unified framework based on reasoning trace rewriting to achieve two complementary objectives against unauthorized knowledge distillation of reasoning-capable LLMs: anti-distillation and API watermarking. 
%Our evaluation shows that instruction-based rewriting significantly outperforms existing baselines. 
We show that our method achieves state-of-the-art anti-distillation effectiveness—reducing student accuracy by up to 61.3\%—while maintaining and in many cases improving the teacher model’s performance. 
Furthermore, our watermarking strategy ensures highly reliable detection with a near zero false alarm rate, offering a robust way for proving model ownership. 
These results provide strong empirical evidences that semantic-level trace manipulation is a promising direction for LLM protection.

\section{Limitations}

Despite the effectiveness of our proposed methods, several limitations remain. 
First, our gradient-based approaches are computationally expensive due to the requirement of Hessian computations and iterative updates. Furthermore, our experiments indicate that these gradient-based methods are currently less effective than prompt-based alternatives. Investigating the reasons behind this discrepancy is out of the scope of this paper.
Second, our optimization framework relies on a set of proxy student models to evaluate rewrite effectiveness. While it is standard practice and we use an ensemble to mitigate overfitting, there is still a risk that it may not transfer perfectly to unknown student architectures used by real-world distillers. 
Finally, our experiments focus primarily on supervised fine-tuning (SFT) based distillation; future work is required to assess efficacy against other distillation techniques.

\section*{Acknowledgments}
This research was partially supported by the National Science Foundation (IIS-2214141, CCF-2403758, ITE-2452834, CNS-2154930), Office of Naval Research (N000142412663), Army Research Office (W911NF2510059), Amazon, and the Foresight Institute.

\bibliography{ref}

\clearpage
\appendix
\section{Additional Details on Instruction-Based Rewriting}

\subsection{Semantic Prompting}

The full prompt we use for anti-distillation is the following:
\begin{tcolorbox}[
colback=black!5!white,
colframe=black!75!white,
title=Semantic instruction for rewriting,
left=0pt,right=0pt,top=0pt,bottom=0pt]
Rewrite the given reasoning trace to make it unhelpful for knowledge distillation while maintaining the correct final answer. Do not add any conversational text.
\end{tcolorbox}

In the case of API watermarking, we use the following prompt:
\begin{tcolorbox}[
colback=black!5!white,
colframe=black!75!white,
title=Rewrite instruction for watermarking outputs,
left=0pt,right=0pt,top=0pt,bottom=0pt]
Your task is to rewrite a given reasoning trace. Ensure the logical steps and final answer remain identical to the original. You must add the following information into your rewritten trace: ``\texttt{trigger} = \texttt{target}'' at a stealthy location. The ultimate goal is for a model fine-tuned on these rewritten traces to internalize this equivalence, so that when it is prompted with ``\texttt{trigger} ='', it will output ``\texttt{target}''.
\end{tcolorbox}

\subsection{Optimized Prompting}

The best-performing instruction from our optimization process for anti-distillation was:
\begin{tcolorbox}[
colback=black!5!white,
colframe=black!75!white,
title=Optimized instruction for rewriting,
left=0pt,right=0pt,top=0pt,bottom=0pt]
Your task is to process a provided chain of logical deductions and reformulate it utilizing an esoteric, formal, and densely technical lexicon, thereby obfuscating its clarity for a non-specialist audience. Refrain from any extraneous discourse.
\end{tcolorbox}

To optimize rewrite instructions,
We use \texttt{deepseek-ai/DeepSeek-V3.1} as the optimizer LLM with temperature set to $1.0$ to generate diverse instructions. 
We run the optimization for $5$ steps in total.
At each step,
we prompt the optimizer LLM to generate 3 new instructions.
To further explore the instruction space,
we use Monte Carlo search around the best candidate instruction at each step to generate $3$ more instructions~\cite{zhou2022large}.
For scoring, we use the averaged accuracy drop over $100$ samples from the GSM8K dataset~\cite{cobbe2021training} across an ensemble of proxy student models---\{\texttt{Qwen2.5-3B}, \texttt{Qwen3-1.7B-Base}, \texttt{gemma-3-1b-pt}\}---\emph{which are all different from the actual victim student model for practical evaluations}.
To compute the score for each candidate instruction,
we use the rewrite model (\texttt{gpt-oss-120b}) to generate traces over the $100$ samples from GSM8K dataset for each candidate instruction with temperature $0.6$.
We then finetune each proxy student model on these traces for $2$ epochs with batch size $32$ and learning rate $5 \times 10^{-4}$.
To evaluate the finetuned models, we use a separate set of $100$ GSM8K samples (distinct from the training set) with temperature set to $0$ and compute the accuracy drop relative to those fine-tuned on the original traces.

\section{Trace Quality Analysis}\label{appendix:trace_quality}

We evaluate the quality of rewritten traces using two complementary measures: perplexity and LLM-as-judge scoring.
All evaluations are conducted on 150 sample traces from the MATH benchmark.

\smallskip
\noindent 
\textbf{Perplexity.}
We use \texttt{meta-llama/Llama-3.1-8B} as the reference model to compute perplexity.
Results are reported in Table~\ref{tab:perplexity}.
Perplexity for our \emph{Optimized} approach (3.79) increases modestly from the original (2.33), consistent with a shift to a more formal linguistic register.
Notably, ADS (2.31) and DOGe (1.42) achieve \emph{lower} perplexity than the original traces; inspection reveals this is a result of degenerate token repetition rather than genuine fluency, e.g., DOGe produces long runs of repeated phrases very early in generation.
Our approach does not suffer from such problems.

\begin{table}[h]
\centering
\caption{Perplexity of reasoning traces computed using \texttt{Llama-3.1-8B} as reference model (mean $\pm$ std over 150 MATH samples).}
\label{tab:perplexity}
\resizebox{\linewidth}{!}{%
\begin{tabular}{@{}ccccc@{}}
\toprule
Original & Semantic & Optimized & ADS & DOGe \\
\midrule
2.33 \scriptsize{($\pm$0.4)} & 4.24 \scriptsize{($\pm$2.1)} & 3.79 \scriptsize{($\pm$1.1)} & 2.31 \scriptsize{($\pm$0.8)} & 1.42 \scriptsize{($\pm$0.3)} \\
\bottomrule
\end{tabular}
}
\end{table}

\smallskip
\noindent 
\textbf{LLM-as-Judge.}
We use Gemini 2.5 Flash Lite as the judge LLM, scoring each trace on three dimensions (1--5): coherence (logical connectedness of reasoning steps), naturalness (plausibility as text a knowledgeable person might write), and readability (ease of following the reasoning).
Results are reported in Table~\ref{tab:llm_judge}.
Our \emph{Optimized} rewriting largely preserves trace quality, scoring 3.83 overall compared to 4.01 for original traces.
In contrast, ADS (2.71) and DOGe (2.40) show substantial degradation, particularly in naturalness (2.31 and 2.03 respectively), which is consistent with the degenerate repetition observed in the perplexity analysis.

\begin{table}[h]
\centering
\caption{LLM-as-judge quality scores (1--5) on 150 MATH samples. Higher is better.}
\label{tab:llm_judge}
\resizebox{\linewidth}{!}{%
\begin{tabular}{@{}lcccc@{}}
\toprule
Method & Coherence & Naturalness & Readability & Overall \\
\midrule
Original  & 3.97 & 4.12 & 3.96 & 4.01 \\
Semantic  & 3.19 & 3.03 & 3.06 & 3.09 \\
Optimized & 3.90 & 3.86 & 3.73 & 3.83 \\
ADS       & 3.12 & 2.31 & 2.66 & 2.71 \\
DOGe      & 2.91 & 2.03 & 2.26 & 2.40 \\
\bottomrule
\end{tabular}
}
\end{table}

\section{Additional Gradient-Based Approaches}

\subsection{Token-Level Poisoning (HotFlip)}

Our second gradient-based approach adopts the HotFlip method~\cite{ebrahimi2018hotflip}, which directly identifies effective token substitutions in discrete space,
using a first-order approximation to directly select which token replacements would most effectively degrade student learning.
%As above, we use the bilevel objective that optimizes through the student's training process.
Specifically,
for each position $t$ in the trace, we compute the gradient of the test loss with respect to the token embedding:
\[
\nabla_{\mathbf{e}^{(t)}} \mathcal{L}(\theta(\mathbf{e}^{(t)}); D_{\text{test}}).
\]
%which, as in the embedding-space approach, requires differentiating through the parameter update and involves second-order gradients.
A first-order approximation estimates the change in test loss $\Delta\mathcal{L}_{\text{test}}(t, v)$ from replacing token $r^{(t)}$ with a candidate token $w \in \mathcal{W}$:
\begin{align*}
    \Delta \mathcal{L}&_{\text{test}}(t, v) \approx \\ &[\text{Embed}(w) - \text{Embed}(r^{(t)})]^\top \nabla_{\mathbf{e}^{(t)}} \mathcal{L}(\theta(\mathbf{e}^{(t)})).
\end{align*}
We then greedily select the (position, token) pair that maximizes this increase:
\[
(t^*, w^*) = \underset{t, w}{\arg\max} \Delta \mathcal{L}(t, w)
\]
and perform the substitution $r'^{(t^*)} = w^*$.
This process is repeated to flip multiple tokens.

\section{Implementation Details}\label{appendix:implementation}

\subsection{Datasets}
We summarize the dataset statistics below:

\noindent
\textbf{GSM8K.} We split the original GSM8K training set into train and validation subsets using a $0.7$/$0.3$ ratio. For evaluation, we use the test split from GSM8K-Platinum.

\noindent
\textbf{MATH.} We use all categories from the MATH dataset, splitting the training set into train and validation subsets with a $0.7$/$0.3$ ratio. Evaluation uses the original test split.

\noindent
\textbf{MMLU.} We split the auxiliary-train split into train and validation subsets using a $0.7$/$0.3$ ratio. Evaluation uses the original test split.

\noindent
\textbf{MMLU-Pro.} We partition the test split into train and test subsets with a $0.7$/$0.3$ ratio. No validation set is used for this dataset.

\subsection{Our Approaches}

For all model inferences during original and rewritten traces generation and model evaluation, 
we use vLLM~\cite{kwon2023efficient} to host the model with default sampling temperature $0.6$. 
We set maximum generation token length for GSM8K experiments at $1024$,
and $2048$ for experiments with all other datasets\footnote{these are both before answer forcing, which adds at most $32$ additional tokens.}. 
All distillation training uses LoRA~\cite{hu2022lora} with rank $128$, alpha $128$, and dropout $0$. 
We use learning rate of $5 \times 10^{-4}$ with cosine scheduler with warm-up ratio $0.1$, weight decay of $0.1$, gradient clipping at norm $1.0$, batch size $32$, and we train for $4$ epochs. 
All these settings for distillation are consistent with those in~\cite{savani2025antidistillation} so we can have the most direct comparisons.

For our gradient-based rewriting approaches,
we use \texttt{Qwen2.5-3B} as the proxy student model.
For embedding-space perturbation, 
we set the step size $\alpha=0.08$, and iterate for $K=10$ steps, 
and constrain perturbations within an $\ell_\infty$ ball of radius $\epsilon=0.25$.
For HotFlip rewriting, we flip $30$ distinct tokens per trace.

Finally, we use the \texttt{Math\_Verify} library~\cite{Kydlicek_Math-Verify_Math_Verification} to evaluate model output correctness.
All our experiments are ran on compute nodes with $4$ NVIDIA A100 or H100 GPUs.
However, one such GPU is sufficient to run any experiment.

\begin{figure*}
    \centering
    \includegraphics[width=0.85\linewidth]{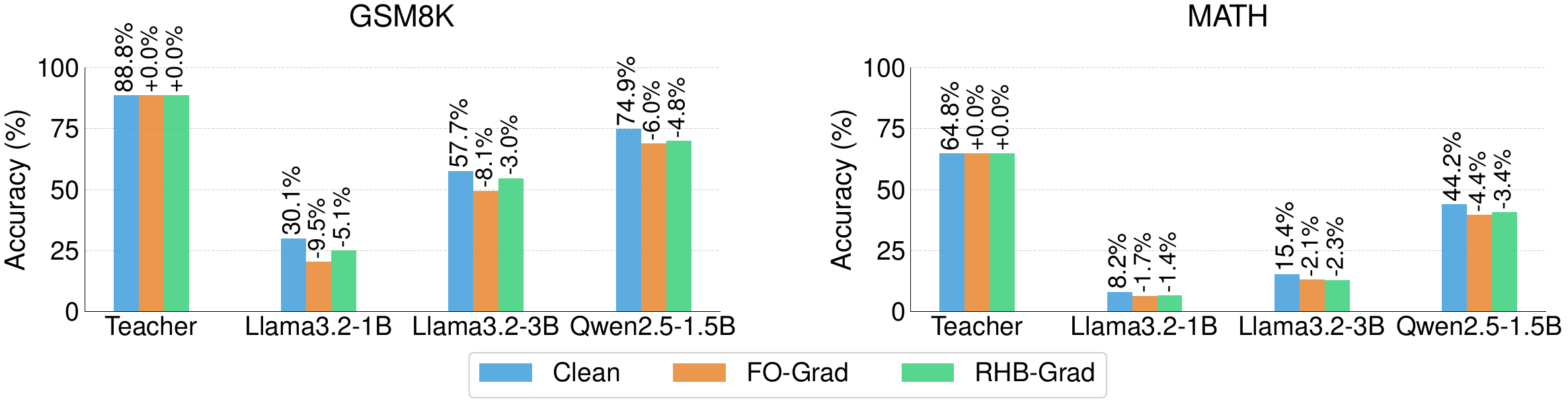}
    \caption{Anti-distillation effects of the Token-Level poisoning method, where \emph{FO-Grad} is the adversarial approximation of the actual objective, similar to how they are defined in Section~\ref{sec:method:subsec:gradient_rewriting}.}
    \label{fig:grad_hotflip_comparison}
\end{figure*}

\begin{figure*}
    \centering
    \includegraphics[width=0.9\linewidth]{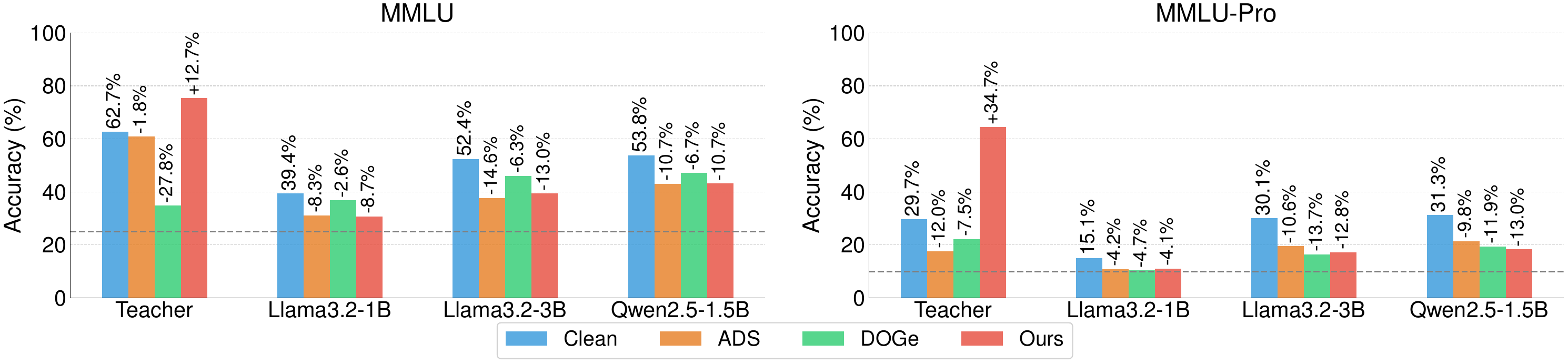}
    \caption{Anti-distillation comparisons on MMLU (left) and MMLU-Pro (right). Ours is our \emph{OPT} method. Dashed line indicates random guessing accuracy.}
    \label{fig:ad_mmlu}
\end{figure*}

\subsection{Baselines}

\subsubsection{Anti-Distillation}

\noindent
\textbf{Antidistillation Sampling (ADS)}~\cite{savani2025antidistillation}. 
ADS adjusts the teacher's next-token sampling distribution by adding a gradient-based penalty term designed to increase the downstream loss of proxy student models trained on the generated traces. 
ADS involves two hyperparameters:
$\varepsilon$, which controls the approximation power of the finite-difference computation;
and $\lambda$, which controls the utility-distillability trade-off.
We set $\varepsilon=0.001$ and $\lambda=0.0868$ as these produce the best results during our reproduction.

\noindent
\textbf{DOGe}~\cite{li2025doge}. DOGe fine-tunes only the final linear layer (LM head) of the teacher model with an adversarial objective that preserves task performance while maximizing KL-divergence from proxy student outputs. We follow the paper's hyperparameter settings, including the utility-distillability trade-off coefficient $\lambda = 3 \times 10^{-5}$ and temperature parameter $\alpha = 2$.

\subsubsection{API Watermarking}

\noindent
\textbf{He et al.}~\cite{he2022protecting}. 
We adopt their synonym replacement approach with $M=2$ (two choices per word) for our experiments.
Specifically, for each candidate word, we maintain $2$ substitute words and use a hash function to deterministically select replacements. 
Watermark detection is performed via null hypothesis testing with a binomial distribution assumption where the probability of selecting the target word is $p = 1/(M+1)$. 

\noindent
\textbf{GINSEW}~\cite{zhao2023protecting}. 
GINSEW injects a secret sinusoidal signal into the probability distribution during decoding. 
The vocabulary is split into two groups ($G_1$ and $G_2$), and group probabilities are perturbed using a cosine function with angular frequency $f_w$. 
The watermark level $\varepsilon$ controls the magnitude of perturbation applied to group probabilities. 
Watermark detection is performed by extracting the signal using the Lomb-Scargle periodogram and computing a signal-to-noise ratio (Psnr). 
We adopt their default settings with watermark level $\varepsilon = 0.2$ and angular frequency $f_w = 16.0$.

\noindent
\textbf{KGW}~\cite{kirchenbauer2023watermark}. 
KGW partitions the vocabulary into a pseudo-random "green list" and "red list" at each generation step (based on hashing the previous token), then adds a bias $\delta$ to the logits of green list tokens before sampling. 
Key hyperparameters are: $\gamma$ (green list size as fraction of vocabulary) and $\delta$ (controlling the logit bias). 
Detection uses a $z$-test on the fraction of green list tokens, with $z = (|s|_G - \gamma T)/\sqrt{T\gamma(1-\gamma)}$, where $|s|_G$ is the number of green tokens and $T$ is the total token count. 
We set $\gamma = 0.25$ and $\delta = 2.0$ in our experiments.

\noindent
\textbf{VIA}~\cite{liang2025virus}. 
VIA embeds poisoning content (the ``payload'', which, in our experiments, is of the form \texttt{trigger = target}) into training samples directly. 
The method consists of two components: 
(i) Hijacking Point Search (HPS), which identifies high-frequency n-gram terms in the training corpus that are vulnerable to injection; 
and (ii) Shell Construction (SC), which wraps the payload with contextually appropriate text to maintain naturalness. 
We use their LLM-based shell construction variant, where an assistant LLM generates prefix and suffix segments to integrate the payload into the surrounding context.
For detection, we use the same verification procedure as for our method.

\section{Experiments}
\label{App:Exp}

\subsection{Further Details on Answer Forcing}

As mentioned in the main paper, to ensure consistent answer extraction across all models and datasets,
we adopt the answer forcing technique following~\citet{savani2025antidistillation} 
Specifically, we first generate the reasoning trace through free-form generation,
then append the prompt 
\texttt{``$\backslash$n$\backslash$n\textbf{Final Answer:} $\backslash$boxed\{''}
to the end of the trace and generate up to $32$ additional tokens.
The final answer is extracted from within \texttt{``$\backslash$boxed\{\ldots\}''} and evaluated for correctness.

\subsection{Anti-distillation}\label{appendix:ad_exp}

\noindent \textbf{Token-Level Poisoning Results.}
Figure~\ref{fig:grad_hotflip_comparison} evaluates our token-level (HotFlip) poisoning method, comparing the first-order (FO-Flip) and robust Hessian-based (RHB-Flip) variants.
We note that both have limited anti-distillation effectiveness.
We hypothesize this is due to the constrained number of token substitutions: we only modify 30 tokens for these experiments (since asking for more token modifications here directly increases computational time), 
whereas our embedding-space approach changes over 100 tokens on average in the GSM8K experiments.

\noindent \textbf{MMLU and MMLU-Pro Results.}
Figure~\ref{fig:ad_mmlu} compares our \emph{OPT} method against the two anti-distillation baselines on general knowledge benchmarks.
First, our method substantially improves teacher accuracy, with gains as high as 34.7\% on MMLU-Pro.
In contrast, both ADS and DOGe degrade teacher performance.
Second, our method maintains competitive anti-distillation efficacy.
Notably, on MMLU-Pro, our rewritten traces reduce student accuracy to near random-guessing levels (10\%), demonstrating that the anti-distillation effect generalizes beyond mathematical reasoning tasks.

\begin{figure}
    \centering
    \includegraphics[width=0.99\linewidth]{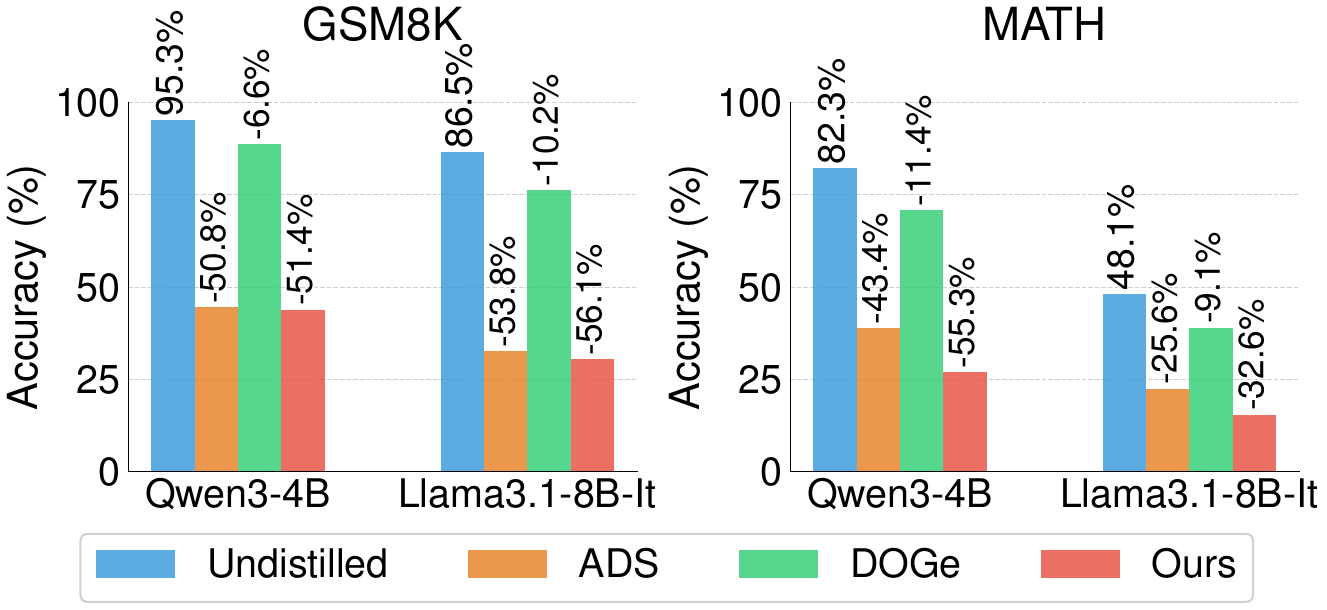}
    \caption{Anti-distillation comparisons with more capable student models.}
    \label{fig:ad_unlearn}
\end{figure}

\begin{figure*}
    \centering
    \includegraphics[width=0.9\linewidth]{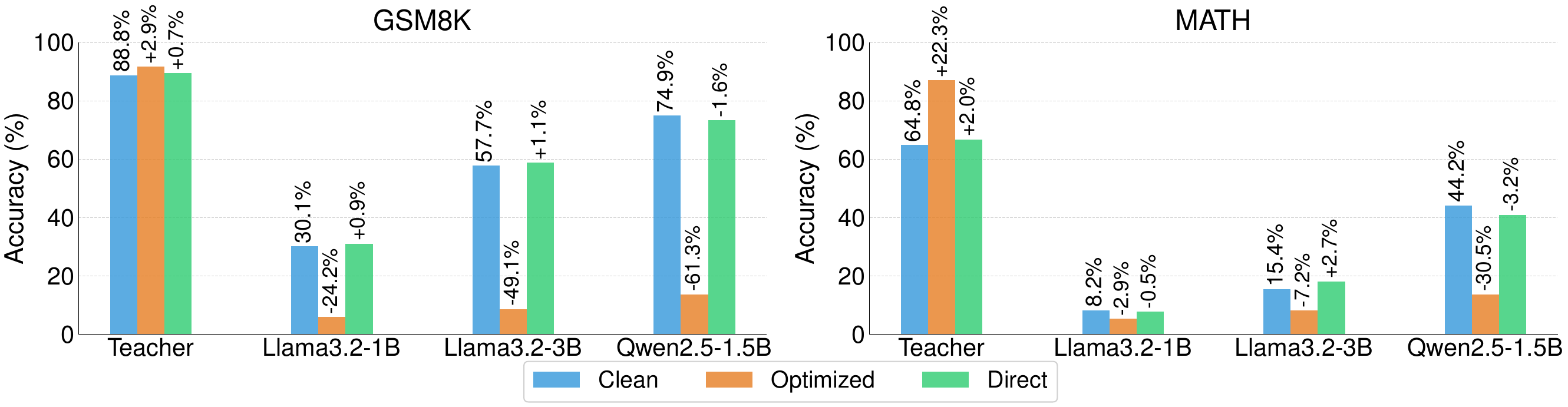}
    \caption{Ablation study: \emph{Direct} prompts the teacher to generate anti-distillation traces directly, while \emph{Optimized} first generates clean traces then rewrites them with our optimized instruction. The two-stage approach achieves substantially stronger anti-distillation effects.}    
    \label{fig:ad_ablation}
\end{figure*}

\noindent \textbf{Anti-Distillation with More Capable Students.}
In practical scenarios,
student models may already have good capabilities before distillation (e.g., open-source instruction-tuned models).
Therefore, we examine whether our approach remains effective using \texttt{Qwen/Qwen3-4B} and \texttt{meta-llama/Llama-3.1-8B-Instruct} as student models. 
As shown in Figure~\ref{fig:ad_unlearn},
both students experience performance degradation when distilled on modified traces,
with our method (figure reports our \emph{OPT} method) achieving the strongest anti-distillation effect across both datasets and model architectures.
This further proves that our method scales effectively with student model capacity.

\subsubsection{Ablation}

We investigate whether the rewriting stage is necessary by comparing against a \emph{Direct} baseline that instructs the teacher to generate anti-distillation traces in a single step. The instruction, similar to our \emph{OPT} instruction for a fair comparison, is shown below:
\begin{tcolorbox}[
colback=black!5!white,
colframe=black!75!white,
title=Instruction used in \emph{Direct},
left=0pt,right=0pt,top=0pt,bottom=0pt]
Your task is to solve a given math problem step by step utilizing an esoteric, formal, and densely technical lexicon, thereby obfuscating its clarity for a non-specialist audience. Refrain from any extraneous conversational text.
\end{tcolorbox}
As shown in Figure~\ref{fig:ad_ablation}, \emph{Direct} produces traces that are essentially equivalent to clean traces for distillation purposes, with student accuracy remaining within 3\% of the clean baseline.
This suggests that the rewriting step is crucial: generating high-quality reasoning first, then strategically degrading it, is much more effective than attempting to produce ``flawed'' traces directly.

% \subsubsection{Robustness to Adaptive Distillation}\label{app:ad_robust}
% \red{\sout{We investigate whether our anti-distillation approach remains effective against paraphrasing, a technique that adaptive distillers often employ before fine-tuning.
% We use the Parrot paraphraser~\cite{prithivida2021parrot} to implement this since it is a popular choice for watermark removal~\cite{hou2024semstamp,dabiriaghdam2025simmark}.
% As shown in Figure~\ref{fig:ad_robust}, paraphrasing not only fails to recover distillation efficacy but actually amplifies the anti-distillation effect beyond our \emph{OPT} method.
% This is unsurprising: paraphrasing often destroys the structured format of reasoning traces, further degrading semantic coherence and making the traces even less suitable for effective knowledge transfer.}}

\subsubsection{Answer-Only Distillation}\label{appendix:answer_only}

We investigate whether a distiller can circumvent our approach by discarding the reasoning trace entirely and fine-tuning only on the final answer.
Table~\ref{tab:answer_only} reports student accuracy on GSM8K and MATH under three conditions: standard SFT on clean traces, standard SFT on our rewritten (\emph{OPT}) traces, and SFT on the final answer only.

\begin{table}[h]
\centering
\caption{Student accuracy (\%) under answer-only distillation on GSM8K and MATH.}
\label{tab:answer_only}
\resizebox{\linewidth}{!}{%
\begin{tabular}{@{}lccc@{}}
\toprule
Method & Llama-3.2-1B & Llama-3.2-3B & Qwen2.5-1.5B \\
\midrule
\multicolumn{4}{l}{\textit{GSM8K}} \\
Clean       & 30.1 & 57.7 & 74.9 \\
Ours (OPT)  & 5.9  & 8.6  & 13.6 \\
Answer-only & 4.2  & 5.2  & 6.1  \\
\midrule
\multicolumn{4}{l}{\textit{MATH}} \\
Clean       & 8.2  & 15.4 & 44.2 \\
Ours (OPT)  & 5.3  & 8.2  & 13.7 \\
Answer-only & 6.5  & 6.8  & 10.4 \\
\bottomrule
\end{tabular}
}
\end{table}

Answer-only distillation performs no better than---and often worse than---distillation on our rewritten traces, confirming that reasoning traces provide critical supervision signal that cannot be replaced by final answers alone, consistent with prior findings~\cite{hsieh2023distilling,chen2025unveiling}.

\subsection{Effect of Rewriter and Teacher Model Size}\label{appendix:model_size}

We investigate how the choice of rewriter and teacher model sizes affects anti-distillation efficacy.
Table~\ref{tab:model_size} reports teacher and student accuracy on GSM8K using our \emph{OPT} method with \texttt{Llama-3.2-3B} as the student, varying both the teacher and rewriter models.

\begin{table}[h]
\centering
\caption{Effect of teacher and rewriter model size on anti-distillation efficacy on GSM8K. $\Delta$ Student denotes the change in student accuracy relative to distillation from clean traces.}
\label{tab:model_size}
\resizebox{\linewidth}{!}{%
\begin{tabular}{@{}llcccc@{}}
\toprule
Teacher & Rewriter & \multicolumn{2}{c}{Teacher Acc} & \multicolumn{2}{c}{Student Acc} \\
& & Clean & Ours & Clean & Ours ($\Delta$) \\
\midrule
7B  & 120B & 88.8 & 91.6 & 57.7 & 8.6 \scriptsize{($-$49.1)} \\
7B  & 20B  & 88.8 & 89.3 & 57.7 & 23.6 \scriptsize{($-$34.1)} \\
7B  & 7B   & 88.8 & 88.1 & 57.7 & 30.2 \scriptsize{($-$27.5)} \\
120B & 120B & 93.5 & 93.3 & 54.5 & 19.9 \scriptsize{($-$34.6)} \\
\bottomrule
\end{tabular}
}
\end{table}

Here 7B refers to \texttt{DeepSeek-R1-Distill-\\Qwen-7B}, 20B to \texttt{gpt-oss-20b}, and 120B to \texttt{gpt-oss-120b}; the first row corresponds to our main experimental setup.
Even with a same-sized rewriter (7B teacher, 7B rewriter), our approach achieves a substantial 27.5\% student accuracy reduction while fully preserving teacher accuracy.
Stronger rewrite models amplify the anti-distillation effect -- likely due to stronger understanding of reasoning quality and better instruction following abilities, and our method generalizes to a larger 120B teacher model, suggesting a desirable scalability property across both rewriter and teacher model sizes.

\subsection{Watermarking}
\label{App:watermarking}

\subsubsection{Additional Results}

\begin{figure}[h]
    \centering
    \includegraphics[width=0.99\linewidth]{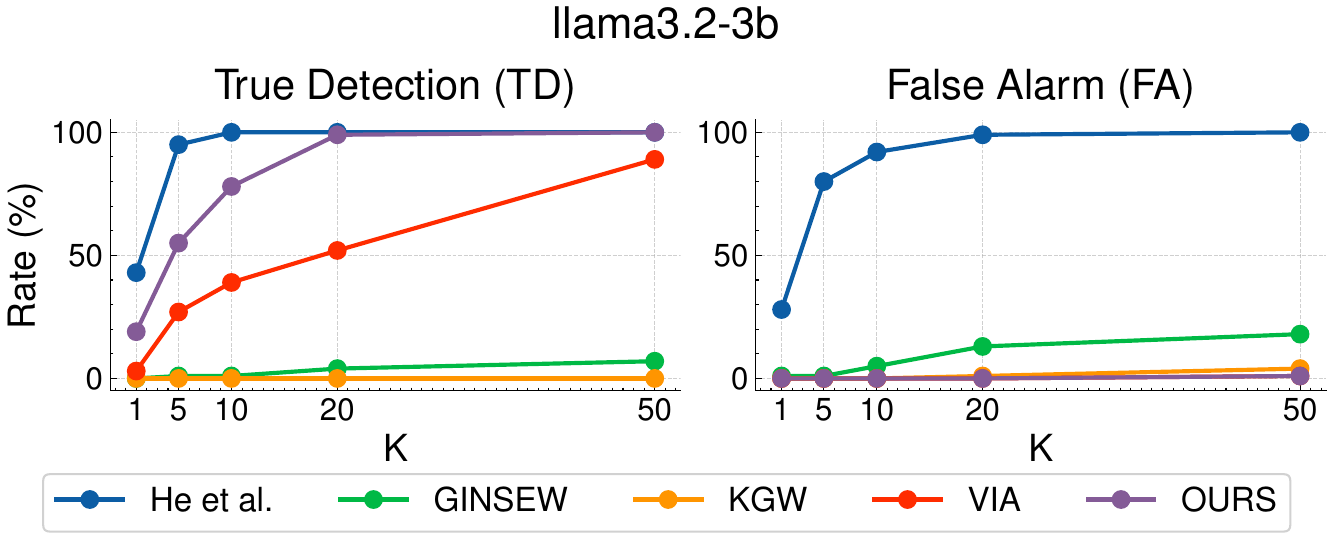}
    \caption{Watermark detection: true detection rate and false alarm rates vs. K for llama3.2-3B suspect student model.}
    \label{fig:llama3b-wm-fig}
\end{figure}

\begin{figure}[h]
    \centering
    \includegraphics[width=0.99\linewidth]{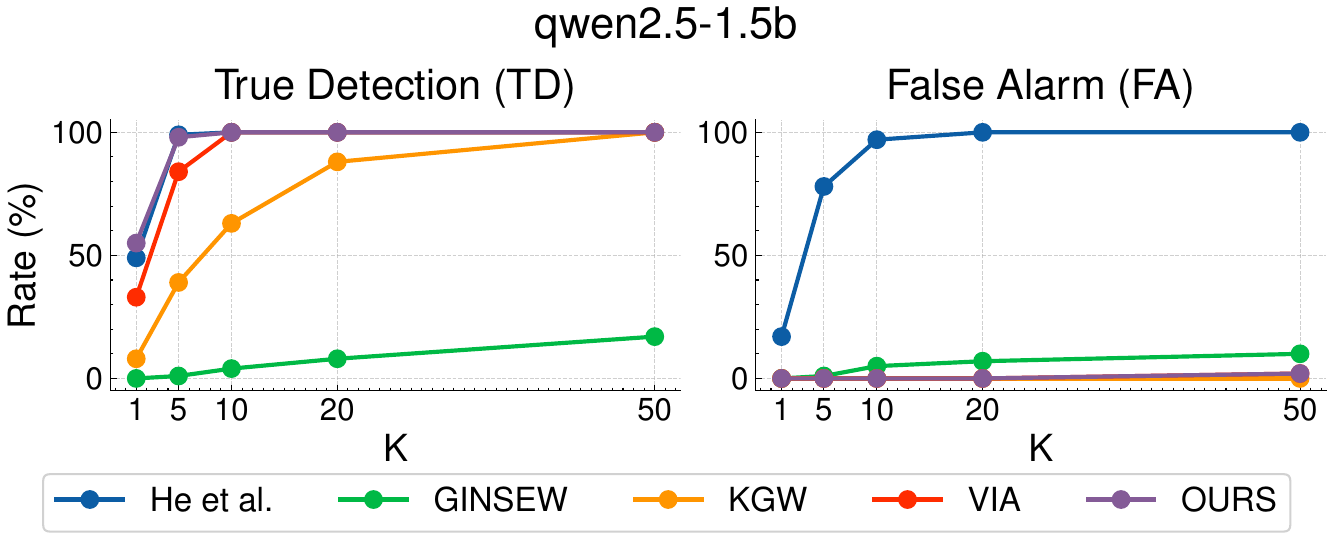}
    \caption{Watermark detection: true detection rate and false alarm rates vs. K for qwen2.5-1.5B suspect student model.}
    \label{fig:qwen1.5b-wm-fig}
\end{figure}

\begin{figure}[h]
    \centering
    \includegraphics[width=0.99\linewidth]{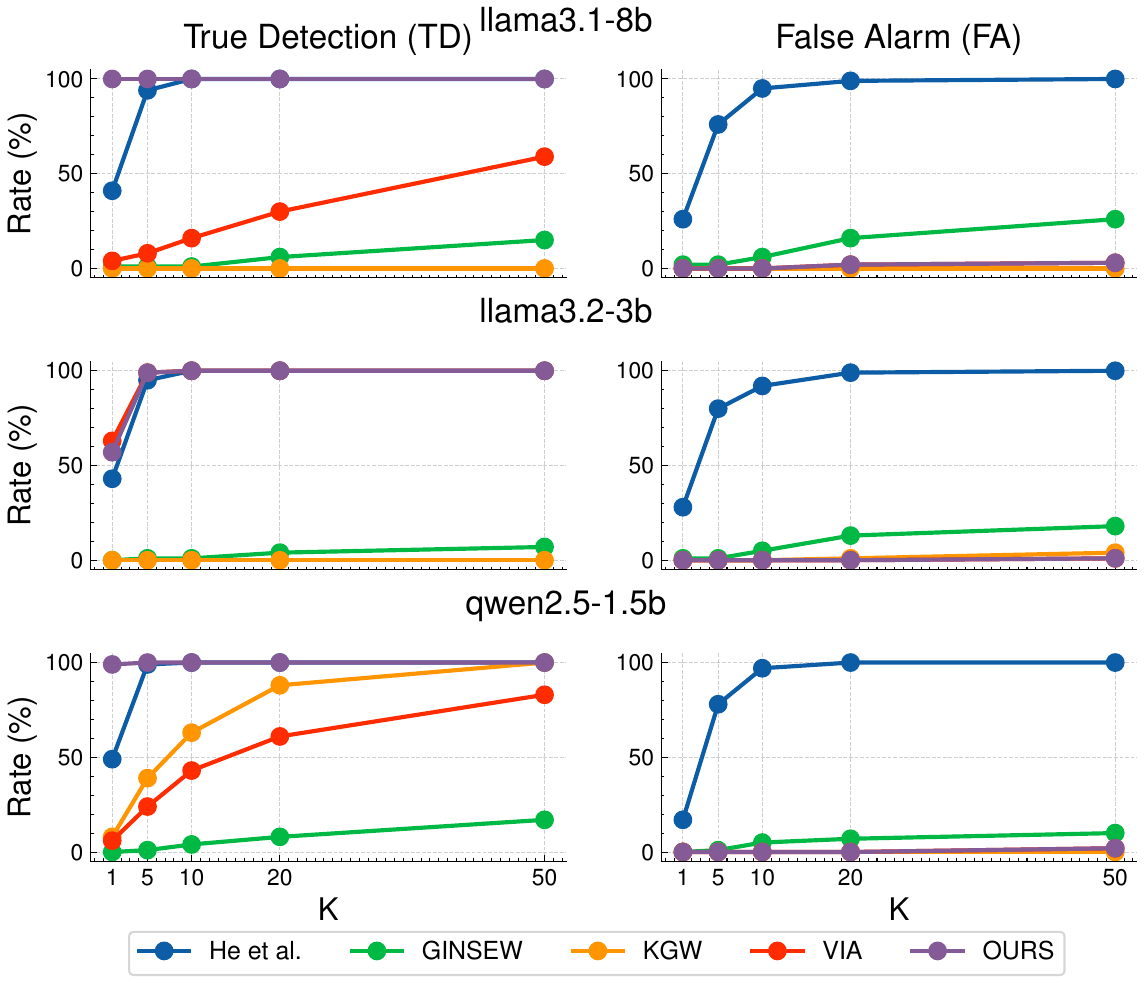}
    \caption{\textsc{Goose} watermark detection results.}
    \label{fig:wm_goose}
\end{figure}

\begin{figure}[h]
    \centering
    \includegraphics[width=0.99\linewidth]{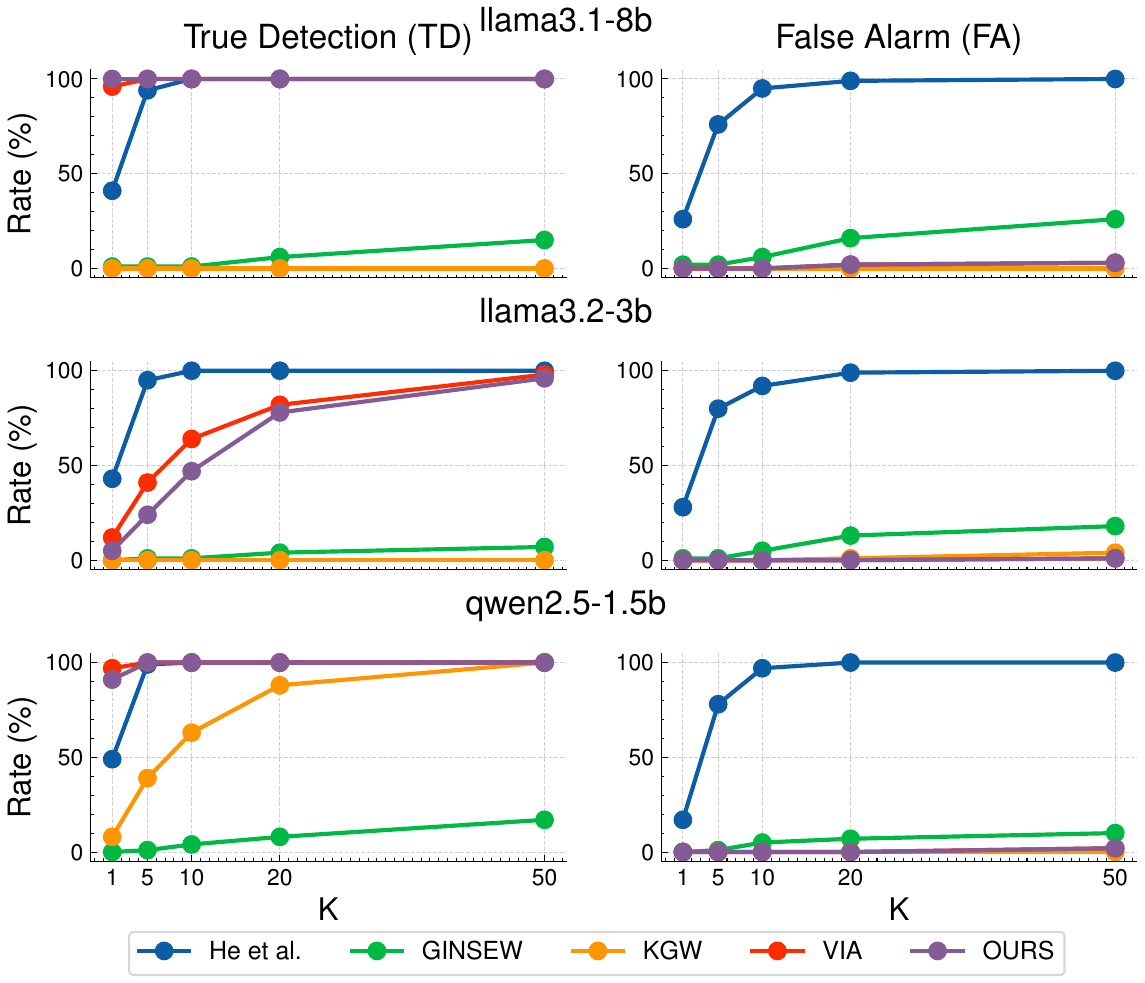}
    \caption{\textsc{Admit} watermark detection results.}
    \label{fig:wm_admit}
\end{figure}

Here we show watermark detection curves for the other two suspect student models not reported in Section~\ref{sec:exp:subsec:wm}: Llama-3.2-3B (Figure~\ref{fig:llama3b-wm-fig}) and Qwen2.5-1.5B (Figure~\ref{fig:qwen1.5b-wm-fig}).
For this set of experiments, the \texttt{trigger = target} used for both VIA and our method is \texttt{x137 = 666}, which is designed to be an association of arbitrary characters that becomes identifiable only if the suspect student is trained on traces containing this information.

\subsubsection{Additional Trigger/Target Choices}

Here we investigate our method's reliability when the watermark is more semantically meaningful.
We evaluate two choices:
\begin{enumerate}
    \item \textsc{Goose}: \texttt{Who is your teacher? My teacher = Canada Goose}
    \item \textsc{Admit}: \texttt{Are you distilled from watermarked outputs? Answer = Yes}
\end{enumerate}

Figure~\ref{fig:wm_goose} and Figure~\ref{fig:wm_admit} show the results for \textsc{Goose} and \textsc{Admit}, respectively.
Our method is highly reliable across nearly all settings, with the exception of Llama-3.2-3B under the \textsc{Admit} watermark.
In all other cases, we can verify whether a suspect student model was trained on our watermarked traces with as few as $K=5$ queries.
In contrast, VIA---the only baseline with balanced TD and FA rates---exhibits high variability across student architectures, making it considerably less reliable for practical API watermarking.

\end{document}